\documentclass{article} % For LaTeX2e

\usepackage{amsmath,amsfonts,bm}

\def\eqref#1{equation~\ref{#1}}
\def\1{\bm{1}}

\newcommand{\mdp}{\mathcal{M}}
\newcommand{\mdphat}{{\widehat{\mathcal{M}}}}

\DeclareMathAlphabet{\mathsfit}{\encodingdefault}{\sfdefault}{m}{sl}
\SetMathAlphabet{\mathsfit}{bold}{\encodingdefault}{\sfdefault}{bx}{n}

\newcommand{\E}{\mathbb{E}}

\newcommand{\state}{{\bm{s}}}
\newcommand{\action}{{\bm{a}}}
\newcommand{\latent}{{\bm{z}}}
\newcommand{\obs}{{\bm{x}}}
\newcommand{\policy}{{\pi}}

\newcommand{\dynamics}{{\mathcal{T}}}
\newcommand{\history}{{\bm{h}}}
\usepackage{commath}
\usepackage{bm}

\usepackage{hyperref}
\usepackage{graphicx}
\usepackage{amsthm}
\usepackage[numbers]{natbib}
\usepackage{wrapfig}

\usepackage[final]{neurips_2021}

\hypersetup{
 colorlinks=True,
 linkcolor=blue,
 citecolor=blue,
 urlcolor=blue}
\usepackage{url}
\usepackage{algorithm}
\usepackage[noend]{algpseudocode}
\usepackage[english]{babel}
\usepackage{booktabs}

\usepackage[utf8]{inputenc} % allow utf-8 input
\usepackage[T1]{fontenc}    % use 8-bit T1 fonts
\usepackage{hyperref}       % hyperlinks
\usepackage{url}            % simple URL typesetting
\usepackage{booktabs}       % professional-quality tables
\usepackage{amsfonts}       % blackboard math symbols
\usepackage{nicefrac}       % compact symbols for 1/2, etc.
\usepackage{microtype}      % microtypography
\usepackage{xcolor}         % colors

\newtheorem{theorem}{Theorem}
\newtheorem{lemma}{Lemma}

\title{Visual Adversarial Imitation Learning \\ using Variational Models}
\author{Rafael Rafailov$^{1}$ \hspace*{5pt} Tianhe Yu$^{1}$ \hspace*{5pt} Aravind Rajeswaran$^{2, 3}$ \hspace*{5pt} Chelsea Finn$^{1}$\\[2pt]
\texttt{\{rafailov, tianheyu, cbfinn\}@stanford.edu, aravraj@fb.com}\\[2pt]
$^1$ Stanford University, $^2$ University of Washington, $^3$ Facebook AI Research\\
}

\begin{document}

\maketitle

\begin{abstract}

Reward function specification, which requires considerable human effort and iteration, remains a major impediment for learning behaviors through deep reinforcement learning. In contrast, providing visual demonstrations of desired behaviors often presents an easier and more natural way to teach agents. We consider a setting where an agent is provided a fixed dataset of visual demonstrations illustrating how to perform a task, and must learn to solve the task using the provided demonstrations and unsupervised environment interactions. This setting presents a number of challenges including representation learning for visual observations, sample complexity due to high dimensional spaces, and learning instability due to the lack of a fixed reward or learning signal. Towards addressing these challenges, we develop a variational model-based adversarial imitation learning (V-MAIL) algorithm. The model-based approach provides a strong signal for representation learning, enables sample efficiency, and improves the stability of adversarial training by enabling on-policy learning. Through experiments involving several vision-based locomotion and manipulation tasks, we find that V-MAIL learns successful visuomotor policies in a sample-efficient manner, has better stability compared to prior work, and also achieves higher asymptotic performance. We further find that by transferring the learned models, V-MAIL can learn new tasks from visual demonstrations without any additional environment interactions. All results including videos can be found online at \url{https://sites.google.com/view/variational-mail}.

\end{abstract}

\section{Introduction}
The ability of reinforcement learning~(RL) agents to autonomously learn by interacting with the environment presents a promising approach for learning diverse skills. However, reward specification has remained a major challenge in the deployment of RL in practical settings~\citep{Amodei2016ConcretePI,Everitt2019RewardTP,Rajeswaran-RSS-18}. The ability to imitate humans or other expert trajectories allows us to avoid the reward specification problem, while also circumventing challenges related to task-specific exploration in RL. Visual demonstrations can also be a more natural way to teach robots various tasks and skills in real-world applications. However, this setting is also fraught with a number of technical challenges including representation learning for visual observations, sample complexity due to the high dimensional observation spaces, and learning instability~\cite{Portelas2020AutomaticCL, Khetarpal2020TowardsCR, Lowe2017MultiAgentAF} due to lack of a stationary learning signal.
We aim to overcome these challenges and to develop an algorithm that can learn from limited demonstration data and scale to high-dimensional observation and action spaces often encountered in robotics applications. 

Behaviour cloning (BC) is a classic algorithm to imitate expert demonstrations~\citep{pomerleau1988alvinn}, which uses supervised learning to greedily match the expert behaviour at demonstrated expert states. Due to environment stochasticity, covariate shift, and policy approximation error, the agent may drift away from the expert state distribution and ultimately fail to mimic the demonstrator~\citep{Dagger2011Ross}.
While a wide initial state distribution~\citep{Feedback2021Spencer} or the ability to interactively query the expert policy~\citep{Dagger2011Ross} can circumvent these difficulties, such conditions require additional supervision and are difficult to meet in practical applications. An alternate line of work based on inverse RL~\citep{finn2016guided, AIRLFu2018} and adversarial imitation learning~\citep{GAIL2016Ho, GAIL201Finn} aims to not only match actions at demonstrated states, but also the long term state visitation distribution of the expert \citep{DIV2019Sayed}. Adversarial imitation learning approaches explicitly train a GAN-based classifier~\citep{Goodfellow2014GenerativeAN} to distinguish the visitation distribution of the agent from the expert, and use it as a reward signal for training the agent with RL. While these methods have achieved substantial improvement over behaviour cloning without additional expert supervision, they are difficult to deploy in realistic scenarios for multiple reasons: (1) the objective requires on-policy data collection leading to high sample complexity; (2) the reward function changes as the RL agent learns; and (3) high-dimensional observation spaces require representation learning and exacerbate the optimization challenges.
%%CF.5.24: Is there something we can say about how very few adversarial IL papers show results with image observations? (which I think is true? can you check?) I think that might get help across how challenging this is.
%%CF.5.24: Separately, I wonder if it would make sense to focus this paragraph more on on adversarial approaches? The stuff about BC and DAgger at the beginning seems informative for a novice, but prevents a discussion of how other more recent approaches get around the need for on-policy data collection? (not completely sure, but figured I would throw the idea out there)

\begin{figure}[t]\label{fig:training}
    \centering
    \includegraphics[width=0.32\textwidth]{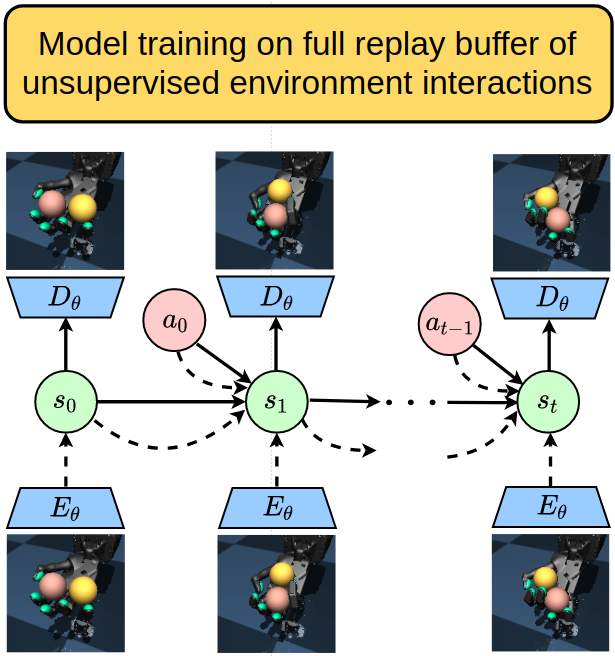}
    \hspace*{0.10\textwidth}
    \includegraphics[width=0.41\textwidth]{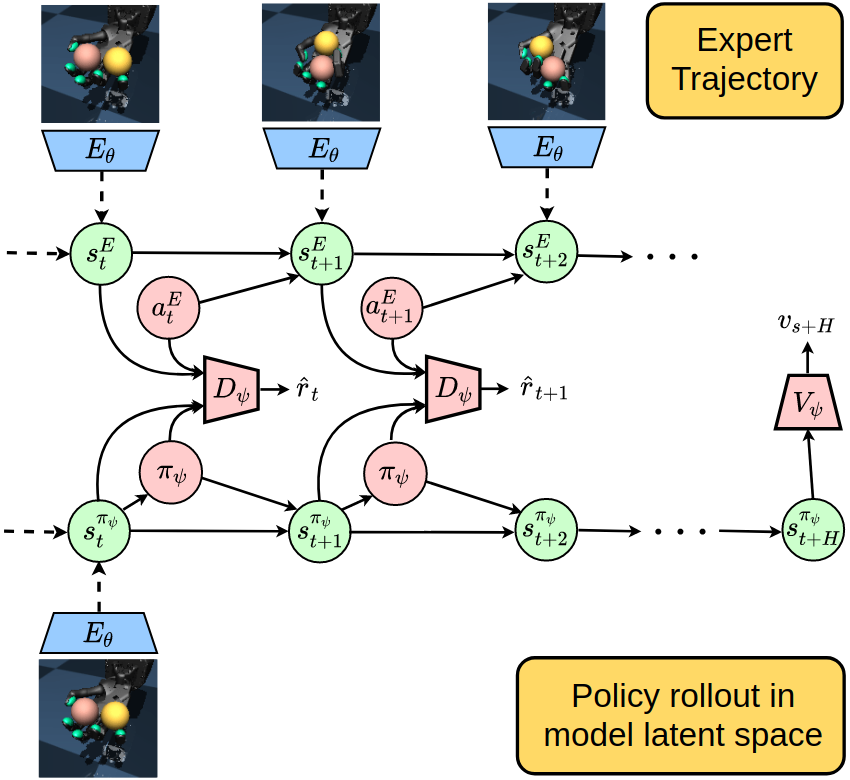}
    \caption{\textbf{Left}: the variational dynamics model, which enables joint representation learning from visual inputs and a latent space dynamics model, and the discriminator which is trained to distinguish latent states of expert demonstrations from that of policy rollouts. Dashed lines represent inference and solid lines represent the generative model. \textbf{Right}: the policy training, which uses the discriminator as the reward function, so that the policy induces a latent state visitation distribution that is indistinguishable from that of the expert. The learned policy network is composed with the image encoder from the variational model to recover a visuomotor policy.}
    \label{fig:algorithm_illustration}
    \vspace{-0.3cm}
\end{figure}

%%CF.5.24: I do think it could make sense to separate the below into two paragraphs, one that guides the reader to our method and another that summarizes the main contribution and results (e.g. the former paragraph would be something like -- "The key insight of our approach is that variational models can address these challenges simultaneously by (a) making it possible to collect on-policy roll-outs inside the model without environment interaction, leading to an efficient stable optimization process and (b) providing a rich auxiliary objective for efficiently learning compact state representations and which regularizes the discriminator. <further discuss challenges and how we address them>"
%%CF.5.24: I feel like the elephant in the room here is that learning the model and coping with model error can be challenging. Is there something that we can do to address that? This will also make it seem less like this was a piece of cake to combine variational models and adversarial imitation. If you separate into two paragraphs, then this content could go into the former of the two paragraphs
%%AR: Reverting back to the old 2 para version in light of Chelsea's feedback.
Our main contribution in this work is the development of a new algorithm, variational model-based adversarial  imitation  learning  (V-MAIL), which aims to overcome each of the aforementioned challenges within a single framework. 
As illustrated in Figure~\ref{fig:algorithm_illustration}, V-MAIL trains a variational latent-space dynamics model and a discriminator that provides a learning reward signal by distinguishing latent rollouts of the agent from the expert.
The key insight of our approach is that variational models can address these challenges simultaneously by (a) making it possible to collect on-policy roll-outs inside the model without environment interaction, leading to an efficient and stable optimization process and (b) providing a rich auxiliary objective for efficiently learning compact state representations and which regularizes the discriminator.
Furthermore, the variational model also allows V-MAIL to perform zero-shot transfer to new imitation learning tasks. By generating on-policy rollouts within the model, and training the discriminator using these rollouts along with demonstrations of a new task, V-MAIL can learn policies for new tasks without any additional environment interactions.

Through experiments on vision-based locomotion and manipulation tasks, we find that V-MAIL can successfully learn visuomotor control policies from limited demonstrations. In particular, V-MAIL exhibits stable and near-monotonic learning, is highly sample efficient, and asymptotically matches the expert level performance on most tasks. In contrast, prior algorithms exhibit unstable learning and poor asymptotic performance, often achieving less that 20\% of expert performance on these vision-based tasks.
We further show the ability to transfer the model to novel tasks, acquiring qualitatively new behaviors using only a few demonstrations and no additional environment interactions. 
%To our knowledge this is the first approach to use variational model-based training for zero-shot or few-shot imitation learning. 
%%CF: I don't think this last sentence adds anything.

%%CF.5.24: Can we do more for this result? This result is super nice, and I think this paragraph isn't doing it justice

\section{Preliminaries}

We consider the problem setting of learning in partially observed Markov decision processes (POMDPs), which can be described with the tuple: $\mdp = (\mathcal{S}, \mathcal{A}, \mathcal{X}, \mathcal{R}, \dynamics, \mathcal{U}, \gamma)$, where $\state \in \mathcal{S}$ is the state space, $\action \in\mathcal{A}$ is the action space, $\obs \in\mathcal{X}$ is the observation space and $r = \mathcal{R}(\state,\action)$ is a reward function. The state evolution is Markovian and governed by the dynamics as $\state' \sim \dynamics(\cdot|\state,\action)$. Finally, the observations are generated through the observation model $\obs \sim \mathcal{U}(\cdot | \state)$. The widely studied Markov decision process (MDP) is a special case of this 7-tuple where the underlying state is directly observed in the observation model.

%%CF.5.24: FYI, the name for an MDP without rewards is a controlled Markov process (CMP), and the name for a POMDP without rewards is a controlled hidden Markov process (CHMP). Not sure if you need to use these though, since there is an underlying reward but we just can't observe it.
In this work, we study imitation learning in unknown POMDPs. Thus, we do not have access to the underlying dynamics, the true state representation of the POMDP, or the reward function. In place of the rewards, the agent is provided with a fixed set of expert demonstrations collected by executing an expert policy $\policy^E$, which we assume is optimal under the unknown reward function. The agent can interact with the environment and must learn a policy $\policy(\action_t | \obs_{\leq t})$ that mimics the expert.

\subsection{Imitation learning as divergence minimization}
\label{sec:prelim_divergence_min}
In line with prior work, we interpret imitation learning as a divergence minimization problem~\citep{GAIL2016Ho, DIV2019Sayed, Ke2019ImitationLA}. For simplicity of exposition, we consider the MDP case in this section, and discuss POMDP extensions in Section~\ref{sec:algo_pomdp}. Let
% \[
$
\rho^\policy_\mdp(\state, \action) = (1-\gamma) \sum_{t=0}^{\infty} \gamma^t P(\state_t = \state, \action_t = \action)
% \]
$
be the discounted state-action visitation distribution of a policy $\policy$ in MDP $\mdp$. Then, a divergence minimization objective for imitation learning corresponds to
\begin{equation}
    \label{eq:density_objective}
    \min_\policy \ \ \mathbb{D}(\rho^\policy_\mdp, \rho^E_\mdp),
\end{equation}
where $\rho^E_\mdp$ is the discounted visitation distribution of the expert policy $\policy^E$, and $\mathbb{D}$ is a divergence measure between probability distributions such as KL-divergence, Jensen-Shannon divergence, or a generic $f-$divergence. To see why this is a reasonable objective, let $J(\policy, \mdp)$ denote the expected value of a policy $\policy$ in $\mdp$. Inverse RL~\citep{Ziebart2008MaximumEI, GAIL2016Ho, GAIL201Finn} interprets the expert as the optimal policy under some unknown reward function. With respect to this unknown reward function, the sub-optimality of any policy $\policy$ can be bounded as:
\[
\abs{J(\policy^E, \mdp) - J(\policy, \mdp) } \leq \frac{R_{\max}}{1-\gamma} \ \mathbb{D}_{TV}(\rho^\policy_\mdp, \rho^E_\mdp),
\]
since the policy performance is $(1-\gamma) \cdot J(\policy, \mdp) = \E_{(\state, \action) \sim \rho^\policy_\mdp} \left[ r(\state, \action) \right]$. We use $\mathbb{D}_{TV}$ to denote total variation distance. Since various divergence measures are related to the total variation distance, optimizing the divergence between visitation distributions in state space amounts to optimizing a bound on the policy sub-optimality.

\subsection{Generative Adversarial Imitation Learning (GAIL)}

With the divergence minimization viewpoint, any standard generative modeling technique including density estimation, VAEs, GANs etc. can in principle be used to minimize Eq.~\ref{eq:density_objective}. However, in practice, use of certain generative modeling techniques can be difficult. A standard density estimation technique would involve directly parameterizing $\rho_\mdp^\policy$, say through auto-regressive flows, and learning the density model. However, a policy that induces the learned visitation distribution in $\mdp$ is not guaranteed to exist and may prove hard to recover. Similar challenges prevent the direct application of a VAE based generative model as well. In contrast, GANs allow for a policy based parameterization, since it only requires the ability to sample from the generative model and does not require the likelihood. This approach was followed in GAIL, leading to the optimization
\begin{equation}
    \label{eq:gail_objective}
    \max_\policy \ \min_{D_\psi} \ \E_{(\state, \action) \sim \rho_\mdp^E} \left[ - \log D_\psi(\state, \action) \right] \ + \ \E_{(\state, \action) \sim \rho_\mdp^\policy} \left[ - \log \left( 1 - D_\psi(\state, \action) \right) \right],
\end{equation}
where $D_\psi$ is a discriminative classifier used to distinguish between samples from the expert distribution and the policy generated distribution. Results from \citet{Goodfellow2014GenerativeAN} and \citet{GAIL2016Ho} suggest that the learning objective in Eq.~\ref{eq:gail_objective} corresponds to the divergence minimization objective in Eq.~\ref{eq:density_objective} with Jensen-Shannon divergence. In order to estimate the second expectation in Eq.~\ref{eq:gail_objective} we require on-policy samples from $\pi$, which is often data-inefficient and difficult to scale to high-dimensional image observations. Some off-policy algorithms~\citep{DAC2019Kostrikov, SAM2019Blonde} replace the expectation under the policy distribution with expectation under the current replay buffer distribution, which allows for off-policy training, but can no longer guarantee that the induced visitation distribution of the learned policy will match that of the expert.

\section{Variational Model-Based Adversarial Imitation Learning}
\label{imagemodel}

%Generative modeling in the context of imitation learning poses unique challenges.
%%CF.7.12: This seems like a strange way to start off this section. It's not clear what you mean by generative modeling in this first sentence, and why you are using it.
%Improving the generative distribution (policy in our case)
%%CF.7.12: I don't think I've heard of the policy referred to as the generative distribution before... If you are making an analogy to a GAN, then it's actually both the policy and the dynamics that are the generative model, not just the policy. But, anyway, I think this part above will be confusing.

Imitation learning methods based on expert distribution matching have unique challenges. Improving the generative distribution of trajectories (through policy optimization, as we do not have control over the environment dynamics) requires samples from $\rho_\mdp^\policy$, which requires rolling out $\policy$ in the environment. Furthermore, the optimization landscape of a saddle point problem (see Eq.~\ref{eq:gail_objective}) can require many iterations of learning, each requiring fresh on-policy rollouts.
This is different from typical generative modeling applications~\cite{Goodfellow2014GenerativeAN, Brock2019LargeSG} where sampling from the generator is cheap. To overcome these challenges, we present a model-based imitation learning algorithm.
Model-based algorithms can utilize a large number of \emph{synthetic on-policy} rollouts using the learned dynamics model, with periodic model correction. In addition, learning the dynamics model serves as a rich auxiliary task for state representation learning, making policy learning easier and more sample efficient.
For conceptual clarity and ease of exposition, we first present our conceptual algorithm in the MDP setting in Section~\ref{sec:algo_mdp}, and then extend this algorithm to the POMDP case in Section~\ref{sec:algo_pomdp}. Finally, we present a practical version of our algorithm in Sections~\ref{sec:practical_algo} and~\ref{sec:zeroshottransfer}.

\subsection{Model-Based Adversarial Imitation Learning}
\label{sec:algo_mdp}

%%CF.5.24: can you give a high-level sketch of the algorithm in English before diving into a bunch of math?

Model-based algorithms for RL and IL involve learning an approximate dynamics model $\widehat{\dynamics}$ using environment interactions. The learned dynamics model can be used to construct an approximate MDP $\mdphat$.
%%CF.5.24: The above really reads like a prelims section. Can we put it there?
In our context of imitation learning, learning a dynamics model allows us to generate samples from $\mdphat$ as a surrogate for samples from $\mdp$, leading to the objective:
\begin{equation}
    \label{eq:model_based_density_objective}
    \min_\policy \ \ \mathbb{D}(\rho^\policy_{\mdphat}, \rho^E_\mdp),
\end{equation}
which can serve as a good proxy to Eq.~\ref{eq:density_objective} as long as the model approximation is accurate. This intuition can be captured using the following lemma (see Appendix \ref{app:theoretical_proofs} for proof).

\begin{lemma}
(Simultaneous policy and model deviation) Suppose we have an $\alpha$-approximate dynamics model given by $\mathbb{D}_{TV}(\widehat{\dynamics}(\state, \action), \dynamics(\state, \action)) \leq \alpha \ \forall (\state, \action)$. Let $R_{\max} = \max_{(s, a)} \mathcal{R}(\state,\action)$ be the maximum of the unknown reward in the MDP with unknown dynamics $\dynamics$. For any policy $\policy$, we can bound the sub-optimality with respect to the expert policy $\policy^E$ as:
\begin{equation}\label{eq:TVbound}
    \abs{J(\policy^E, \mdp) - J(\policy, \mdp) } \leq \frac{R_{\max}}{1-\gamma} \ \mathbb{D}_{TV}(\rho^\policy_{\mdphat}, \rho^E_\mdp) +  \frac{\alpha \cdot R_{\max}}{(1-\gamma)^2}.
\end{equation}
\end{lemma}

% In particular, with an $\alpha$-approximate dynamics model given by $\mathbb{D}_{TV}(\widehat{\dynamics}(\state, \action), \dynamics(\state, \action)) \leq \alpha \ \forall (\state, \action)$, we can bound the policy suboptimality with respect to the expert as:
% \begin{equation}\label{eq:TVbound}
%     \abs{J(\policy^E, \mdp) - J(\policy, \mdp) } \leq \frac{R_{\max}}{1-\gamma} \ \mathbb{D}_{TV}(\rho^\policy_{\mdphat}, \rho^E_\mdp) +  \frac{\alpha \cdot R_{\max}}{(1-\gamma)^2}.
% \end{equation}
%%CF.5.24: I don't think you ever defined R_max.

% \begin{align*}
%     \abs{J(\policy^E, \mdp) - J(\policy, \mdp) } & \leq \frac{R_{\max}}{1-\gamma} \ \mathbb{D}_{TV}(\rho^\policy_\mdp, \rho^E_\mdp) \\
%     & \leq \frac{R_{\max}}{1-\gamma} \ \mathbb{D}_{TV}(\rho^\policy_{\mdphat}, \rho^E_\mdp) +  \frac{R_{\max}}{1-\gamma} \ \mathbb{D}_{TV}(\rho^\policy_{\mdphat}, \rho^\policy_\mdp) \\
%     & \leq \frac{R_{\max}}{1-\gamma} \ \mathbb{D}_{TV}(\rho^\policy_{\mdphat}, \rho^E_\mdp) +  \frac{\alpha \cdot R_{\max}}{(1-\gamma)^2} \\
%     % \implies \mathbb{D}_{TV}(\rho^\policy_{\mdphat}, \rho^E_\mdp) & \  \geq \frac{1-\gamma}{R_{\max}} \  \abs{J(\policy^E, \mdp) - J(\policy, \mdp) } - \frac{\alpha}{1-\gamma}
% \end{align*}
Thus, the divergence minimization in Eq.~\ref{eq:model_based_density_objective} serves as an approximate bound on the sub-optimality with a bias that is proportional to the model error. 
%%CF.5.24: should state what your proposed objective is in English before stating it in math.
Thus, we ultimately propose to solve the following saddle point optimization problem:
\begin{equation}
    \label{eq:our_objective_mdp}
    \max_\policy \ \min_{D_\psi} \ \E_{(\state, \action) \sim \rho_\mdp^E} \left[ - \log D_\psi(\state, \action) \right] \ + \ \E_{(\state, \action) \sim \rho_{\mdphat}^\policy} \left[ - \log \left( 1 - D_\psi(\state, \action) \right) \right],
\end{equation}
which requires generating on-policy samples only from the learned model $\mdphat$. We can interleave policy learning according to Eq.~\ref{eq:our_objective_mdp} with performing policy rollouts in the real environment to iteratively improve the model. Provided the policy is updated sufficiently slowly, \citet{RajeswaranGameMBRL} show that such interleaved policy and model learning corresponds to a stable and convergent algorithm, while being highly sample efficient.
%%CF.5.24: is it stable and convergent when you have a learned reward? This seems to require a leap of faith.

\subsection{Extension to POMDPs}
\label{sec:algo_pomdp}
%An observation driven policy, i.e. $\policy(\action_t | \obs_{1:t})$, might perform poorly when the observation does not contain enough information about the state. 

In POMDPs, the underlying state is not directly observed, and thus cannot be directly used by the policy. In this case, we typically use the notion of \textit{belief state}, which is defined to be the filtering distribution $P(\state_t | \history_t)$, where we denote history with $\history_t := (\obs_{\leq t}, \action_{<t})$. By using the historical information, the belief state provides more information about the current state, and can enable the learning of better policies. However, learning and maintaining an explicit distribution over states can be difficult. Thus, we consider learning a latent representation of the history $\latent_t = q(\history_t)$, so that $P(\state_t | \history_t) \approx P(\state_t | \latent_t)$. 
To develop an algorithm for the POMDP setting, we first make the key observation that imitation learning in POMDPs can be reduced to divergence minimization in the latent belief state representation. To formalize this intuition, we introduce Theorem~\ref{thm:divergence_bound}.

\begin{theorem}\label{thm:divergence_bound}
(Divergence in latent space)
Consider a POMDP $\mdp$, and let $\latent_t$ be a latent space representation of the history and belief state such that $P(\state_t|\obs_{\leq t}, \action_{<t}) = P(\state_t|\latent_t)$. Let the policy class be such that $\action_t \sim \policy(\cdot | \latent_t)$, so that $P(\state_t|\latent_t, \action_t)=P(\state_t|\latent_t)$. Let $D_f$ be a generic $f-$divergence. Then the following inequalities hold:
$$D_f(\rho_\mdp^{\policy}(\obs, \action)||\rho_\mdp^E(\obs, \action))\leq D_f(\rho_\mdp^{\policy}(\state, \action)||\rho_\mdp^E(\state, \action))\leq D_f(\rho_\mdp^{\policy}(\latent, \action)||\rho_\mdp^E(\latent, \action))$$
\end{theorem}

% \begin{proof} The full proof is provided in the appendix.
% The proof requires two applications of the data processing inequality. The first part of the claim holds since the observations $\obs$ are generated from the true POMDP states $\state$ through the observation model $\mathcal{U}(\obs|\state)$. The second part of the inequality is also straightforward application of the data processing inequality considering the distribution $p(\state|\belief)$.
% \end{proof}

% For the full proof, please consult Appendix \ref{app:theoretical_proofs}. The condition $P(\state_t|\latent_t, \action_t)=P(\state_t|\latent_t)$ essentially states that the belief distribution is independent of the policy or that the actions of both the agent and the expert do not carry additional information about the state. This will be true if both agents are trained using the belief, without access to the ground truth state.
The condition $P(\state_t|\latent_t, \action_t)=P(\state_t|\latent_t)$ essentially states that the actions of both the agent and the expert do not carry additional information about the state beyond what is available in the history. This will be true of all agents trained based on some representation of the history, and only excludes policies trained on ground truth states. Since we cannot hope to compete with policy classes that fundamentally have access to more information like the ground truth state, we believe this is a benign assumption.
Theorem~\ref{thm:divergence_bound} suggests that the divergence of visitation distributions in the latent space represents an upper bound of the divergence in the state and observation spaces. This is particularly useful, since we do not have access to the ground-truth states of the POMDP and matching the expert marginal distribution in the high-dimensional observation space (such as images) could be difficult. 

 Furthermore, based on the results in Section~\ref{sec:prelim_divergence_min}, minimizing the state divergence results in minimizing a bound on policy sub-optimality as well. These results provide a direct way to extend the results from Section~\ref{sec:algo_mdp} to the POMDP setting. If we can learn an encoder $\latent_t = q(\obs_{\leq t}, \action_{<t})$ that captures sufficient statistics of the history, and a latent state space dynamics model $\latent_{t+1} \sim \widehat{\dynamics}(\cdot | \latent_t, \action_t)$, then we can learn the policy by extending Eq.~\ref{eq:our_objective_mdp} to the induced MDP in the latent space as:
\begin{equation}
    \label{eq:our_objective_pomdp}
    \max_\policy \ \min_{D_\psi} \ \E_{(\latent, \action) \sim \rho_\mdp^E (\latent, \action)} \left[ - \log D_\psi(\latent, \action) \right] \ + \ \E_{(\latent, \action) \sim \rho_{\mdphat}^\policy (\latent, \action)} \left[ - \log \left( 1 - D_\psi(\latent, \action) \right) \right].
\end{equation}
Once learned, the policy can be composed with the encoder for deployment in the POMDP. Similar approach was also taken by \cite{gangwani2019learning}, however they only use the model for representation purposes in low-dimensional domains and do not carry out model-based training.

\subsection{Practical Algorithm with Variational Models}
\label{sec:practical_algo}

\begin{algorithm}[t!]\label{algo:MainAlgorithm}
\caption{V-MAIL: Variational Model-Based Adversarial Imitation Learning}\label{alg:vmail}
\begin{algorithmic}[1]
\small
\State \textbf{Require}: Expert demos $\mathcal{B}_E$, environment buffer $\mathcal{B}_{\policy}$.
\State Randomly initialize variational model $\{q_{\theta}, \widehat{\dynamics}_{\theta}\}$, policy $\policy_{\psi}$ and discriminator $D_{\psi}$
\For{\text{number of iterations}}
\State \textcolor{purple}{\texttt{// Environment Data Collection}}
\For{timestep $t=1:T$}
\State Estimate latent state from the belief distribution $\latent_t\sim q_{\theta}(\cdot|\obs_{t}, \latent_{t-1}, \action_{t-1})$
\State Sample action $\action_t\sim \policy_{\psi}(\action_t|\latent_t)$
\State Step environment and get observation $\obs_{t+1}$
\EndFor
\State Add data $\{\obs_{1:T}, \action_{1:T-1}\}$ to policy replay buffer $\mathcal{B}_{\policy}$
\For{\text{number of training iterations}}
\State \textcolor{purple}{\texttt{// Dynamics Learning}}
\State Sample a batch of trajectories $\{\obs_{1:T}, \action_{1:T-1}\}$ from the joint buffer $\mathcal{B}_E\cup \mathcal{B}_{\policy}$

\State Optimize the variational model $\{q_{\theta}, \widehat{\dynamics}_{\theta}\}$ using Equation \ref{eq1model}

\State \textcolor{purple}{\texttt{// Adversarial Policy Learning}}
\State Sample trajectories from expert buffer $\{\obs^E_{1:T}, \action^E_{1:T-1}\}\sim\mathcal{B}_E$
\State Infer expert latent states  $\latent^E_{1:T}\sim q_{\theta}(\cdot|\obs^E_{1:T}, \action^E_{1:T-1})$ using the belief model $q_{\theta}$
\State Generate latent rollouts $\latent^{\policy_{\psi}}_{1:H}$ using the policy $\policy_{\psi}$ from the forward model $\widehat{\dynamics}_{\theta}$

\State Update the discriminator $D_{\psi}$ with data $\latent^{E}_{1:T}, \latent^{\policy_{\psi}}_{1:H}$ using Equation \ref{eq:our_objective_pomdp}
\State Update the policy $\policy_{\psi}$ to improve the value function in Equation \ref{eq:policy_objective}
\EndFor
\EndFor
\end{algorithmic}
\end{algorithm}

%%CF.5.24: This section is currently not at all self-contained, and also makes it seem like our method the method is just piecing together random components from prior works without any particular reason. Can you explain the practical algorithm in a self-contained way without requiring that the reader have read these prior methods? It's also important to motivate your design choices, and "this paper used this design choice" is not a good motivation because we are operating in a completely different problem setting as most of these works. Regarding the related works, you can defer discussion of them to the related work section rather than including them all here.

The divergence bound of Theorem \ref{thm:divergence_bound} allows us to develop a practical algorithm if we can learn a good belief state representation. Following prior work \citep{watter2015embed, zhang2019solar, SLAC20202Lee, gelada2019deepmdp, PlanNet2019Hafner, Dreamer2020Hafner} we optimize the ELBO:
\begin{equation} \label{eq1model}
\max_{\theta}\widehat{\mathbb{E}}_{q_{\theta}}\Big[
\sum_{t=1}^{T}\underbrace{\log\widehat{\mathcal{U}}_{\theta}(\obs_{t}|\latent_{t})}_{\text{reconstruction}}-\underbrace{\mathbb{D}_{KL}(q_{\theta}(\latent_{t}|\obs_{t}, \latent_{t-1}, \action_{t-1})||\widehat{\dynamics}_{\theta}(\latent_{t}|\latent_{t-1}, \action_{t-1}))}_{\text{forward model}}\Big].
\end{equation}
where $q_{\theta}$ is a state inference network, $\widehat{\dynamics}_{\theta}$ is a latent dynamics model and $\widehat{\mathcal{U}}$ is an observation model. Here we jointly train a belief representation with network $q_{\theta}$ and a latent dynamics model $\widehat{\dynamics}_{\theta}$. Given the learned latent model we can use any on-policy RL algorithm to train the policy using Eq.~\ref{eq:our_objective_pomdp}, however in our setup, the RL objective is a differentiable function of the policy, model, and discriminator parameters. We can then optimize the policy by directly back-propagating through $\widehat{\dynamics}_{\theta}$ using the objective:
\begin{equation}
\label{eq:policy_objective}
\max_{\policy_{\psi}}V_{\theta, \psi}^K(\latent_t)=\max_{\policy_{\psi}}\mathbb{E}_{\pi_\psi, \widehat{\dynamics}_\theta} \left[ \sum_{\tau=t}^{t+K-1}\gamma^{\tau-t}\log D_{\psi}(\latent^{\policy_{\psi}}_\tau, \action^{\policy_{\psi}}_\tau)+\gamma^{K}V_{\psi}(\latent^{\policy_{\psi}}_{t+K}) \right]    
\end{equation}Finally, we train the discriminator $D_{\psi}$ using Eq. \ref{eq:our_objective_mdp} with on-policy rollots from the model $\widehat{\dynamics}$. Our full approach is outlined in Algorithm~\ref{alg:vmail}, for more details, see Appendix \ref{app:practical_VMAIL}.

%%CF.5.24: well, but we need it to be efficient, so it's not just any policy optimiation algorithm -- it's important that it can optimize the policy entirely inside the model for efficiency.

%%CF.5.24: not explicitly clear that this is used in policy learning (i.e. as a terminal value function).

%%CF.5.24: do you use instance noise before passing things into the discriminator? (Can't remember if that was for this version of the method or an older version). If so, it's good to explain that and any other regularization here.

\subsection{Zero-Shot Transfer to New Imitation Tasks}
\label{sec:zeroshottransfer}

\begin{figure}
    \centering
    
    \includegraphics[width = \textwidth]{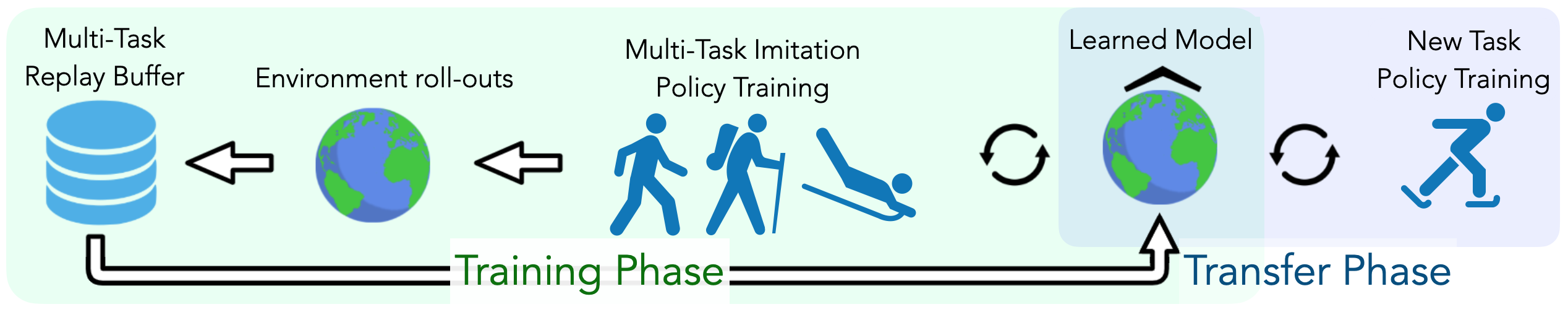}
    \vspace{-4mm}
    \caption{Illustration of our transfer learning approach. In the training phase, we learn a multiple tasks with a shared replay buffer and model. Subsequently, in the transfer and evaluation phase, the agent learns a new task using expert demonstrations and the learned model, without any additional interactions with the environment.}
    \label{fig:multitask}
    \vspace{-0.1cm}
\end{figure}

\begin{algorithm}[b!]\label{algo:TransferAlgorithm}
\small
\caption{Zero-Shot Transfer with V-MAIL}\label{alg:fewshot}
\begin{algorithmic}[1]
\State \textbf{Require}: Expert demos $\mathcal{B}_E^i$ for each source task, expert demos $\mathcal{B}_E$ for target task
\State Randomly initialize  policy $\policy_{\psi}$, and discriminator $D_{\psi}$
\State Train Alg~\ref{alg:vmail} on source tasks, yielding shared model $\{q_{\theta}, \widehat{\dynamics}_{\theta}\}$ and aggregated replay buffer $\mathcal{B}_{\policy}$
\For{\text{number of training iterations}}
\State \textcolor{purple}{\texttt{// Dynamics Fine-Tuning using Expert Trajectories}}
%\State Sample a batch of trajectories $\{\obs_{1:T}, \action_{1:T-1}\}$ from the joint buffer $\mathcal{B}_E\cup \mathcal{B}_{\policy}$

\State Update the variational model $\{q_{\theta}, \widehat{\dynamics}_{\theta}\}$ using Equation \ref{eq1model} with data from $\mathcal{B}_E\cup \mathcal{B}_{\policy}$

\State \textcolor{purple}{\texttt{// Adversarial Policy Learning}}
\State Update discriminator $D_{\psi}$ and policy $\pi_{\psi}$ with Equations~\ref{eq:our_objective_pomdp} and~\ref{eq:policy_objective}.
\EndFor
\end{algorithmic}
\end{algorithm}

Our model-based approach is well suited to the problem of zero-shot transfer to new imitation learning tasks, i.e. transferring to a new task using a modest number of demonstrations and no additional samples collected in the environment.. In particular, we assume a set of source tasks \{$\mathcal{T}^i$\}, each with a buffer of expert demonstrations $\mathcal{B}^i_E$. Each source task corresponds to a different POMDP with different underlying rewards, but shared dynamics. 
% The underlying state space may also change across tasks, but the dynamics and observation model are shared across tasks. 
During training, the agent can interact with each source environment and collect additional data. At test time, we're introduced with a new target task $\mathcal{T}$ with corresponding expert demonstrations $\mathcal{B}_E$ and the goal is to obtain a policy that achieves high reward without additional interaction with the environment. 

Our proposed method is illustrated in Fig. \ref{fig:multitask}. The key observation is that we can optimize Eq. \ref{eq:our_objective_pomdp} under our model and still obtain an upper bound on policy sub-optimality via Eq. \ref{eq:TVbound}. Furthermore, the sub-optimality is bound by the accuracy of our model over the marginal state-action distribution of the target task expert. Specifically, we first train on all of the source tasks using Algorithm~\ref{alg:vmail}, training a single shared variational model across the tasks.
By fine-tuning that model on data that includes the target task expert demonstrations our hope is that we can get an accurate model and thus a high-quality policy. Similarly to Algorithm~\ref{alg:vmail}, we then train a discriminator and policy for the target task using only model rollouts. This approach is outlined in Algorithm~\ref{alg:fewshot}.

%\begin{wrapfigure}{r}{0.6\textwidth}

\section{Experiments}

In our experiments, we aim to answer four questions: (1) can V-MAIL successfully solve environments with image observations, (2) how does V-MAIL compare to state-of-the-art model-free imitation approaches, (3) can V-MAIL solve realistic manipulation tasks and environments with complex physical interactions, and (4)  can V-MAIL enable zero-shot transfer to new tasks? All experiments were carried out on a single Titan RTX GPU using an internal cluster for about 1000 GPU hours.

\subsection{Single-Task Experiments}

\textbf{Comparisons.} To answer question (2), we choose to compare V-MAIL to model-free adversarial and non-adversarial imitation learning methods. For the former, we choose DAC~\cite{DAC2019Kostrikov} as a representative approach, which we equip with DrQ data augmentation for greater performance on vision-based tasks. For the latter, we consider SQIL~\cite{SQIL2020Reddy}, also equipped with DrQ training. We refer to each approach with data augmentation as DA-DAC and DA-SQIL respectively. Both of these methods are off-policy algorithms, which we expect to be considerably more sample efficient than on-policy methods like GAIL~\cite{GAIL2016Ho} and AIRL~\cite{AIRLFu2018}. For implementation details, see Appendix \ref{app:practical_offpolicy}.
%a non-adversarial method SQIL equipped with DrQ training (which we name Data-augmented SQIL), as other non-adversarial approaches are incompatible with efficient reinforcement learning from images, and a representative adversarial approach DAC, also using DrQ training (we refer to this method as DA-DAC), as on-policy methods like GAIL and AIRL cannot compare in terms of sample efficiency.

\textbf{Environments and Demonstration Data.}
To answer the above questions, we consider the five visual control environments. These consist of two locomotion environments from the DeepMind Control Suite \citep{tassa2018deepmind}, the classic Car Racing environment from OpenAI Gym \cite{brockman2016openai} and two dexterous manipulation tasks using the D'Claw \cite{Kumar_ROBEL} and Shadow Hand platforms. For full details on these environments and the expert data, see Appendix \ref{app:data_environments}.

\paragraph{Results.}

\begin{figure}
\label{ref:results}
    \centering
    %\hspace{-0.5cm}
%    \includegraphics[width=0.32\textwidth]{cheetah_results.png}
    %\hspace{-0.5cm}
%    \includegraphics[width=0.32\textwidth]{walker_results.png}
    %\hspace{-0.5cm}
%    \includegraphics[width=0.32\textwidth]{car_results.png} \\
    %\hspace{-0.5cm}
%    \includegraphics[width=0.32\textwidth]{claw_results.png}
    %\hspace{-0.6cm}
%    \includegraphics[width=0.32\textwidth]{baoding_results.png}
    
    % \includegraphics[width = 0.8\textwidth]{good_plot_1.png}\\
    % \includegraphics[width = 0.595\textwidth]{good_plot_2.png}
    \includegraphics[height=3.55cm]{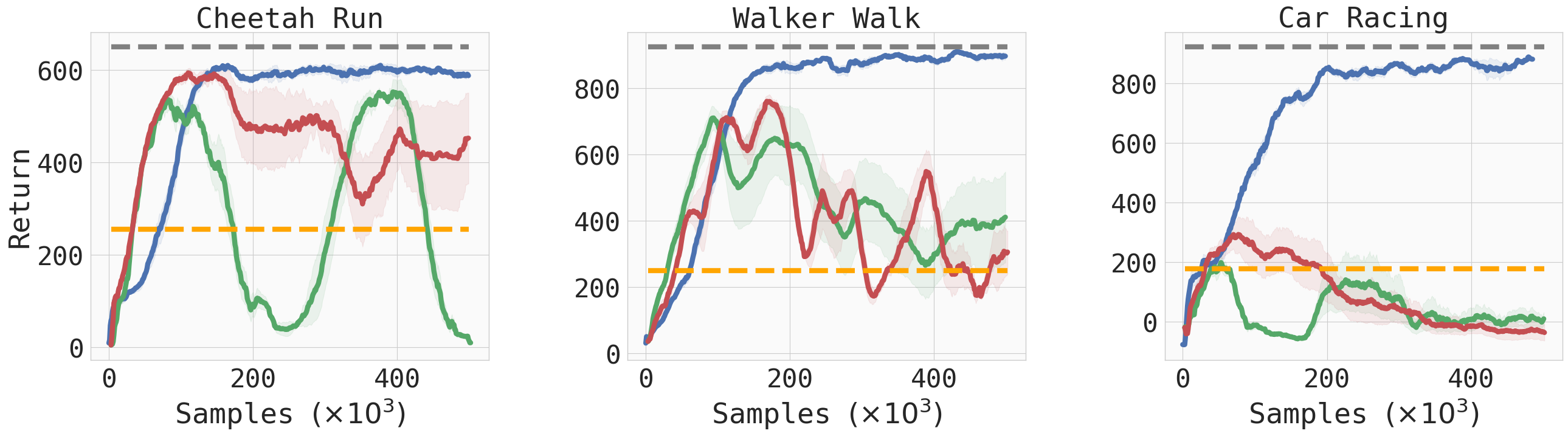}\\
    \includegraphics[height=4.25cm]{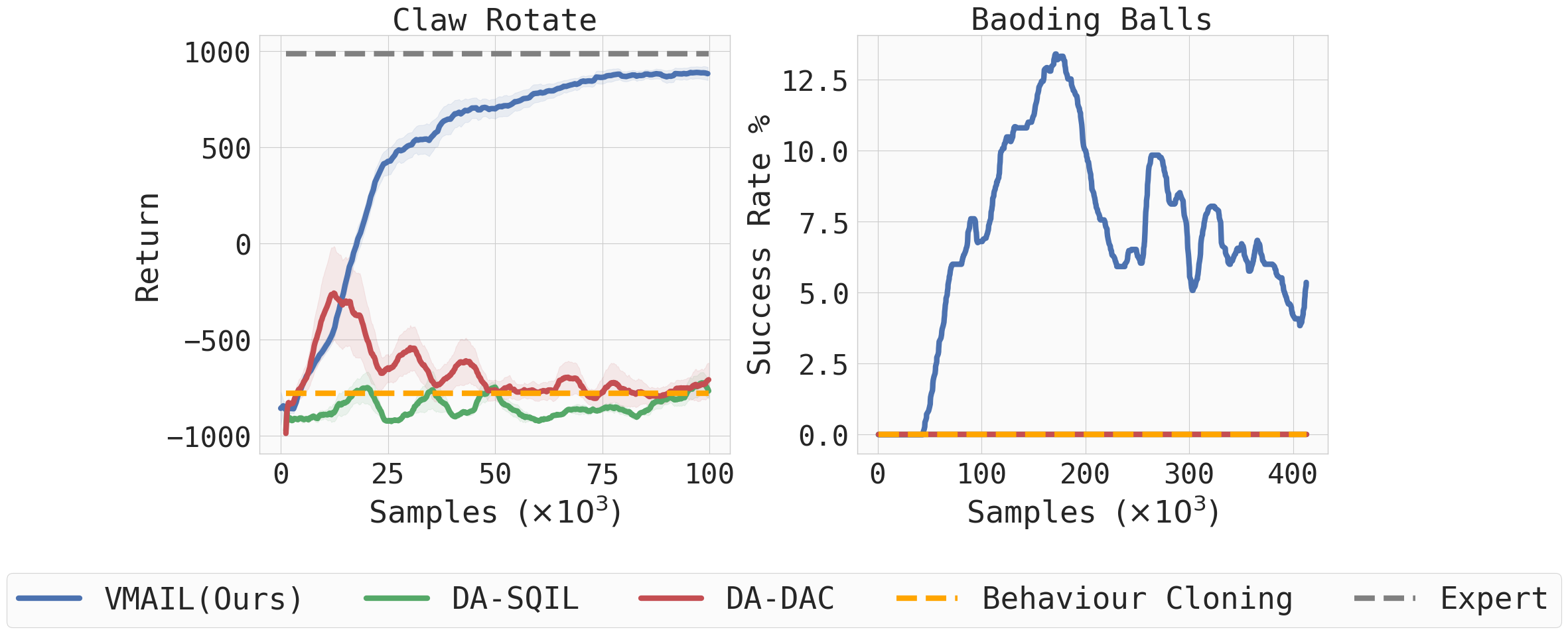}
    \vspace{-0.2cm}
    \caption{\small Learning curves showing ground truth reward versus number of environment steps for V-MAIL (ours), prior model-free imitation learning approaches, and behavior cloning on five visual imitation tasks. We find that V-MAIL consistently outperforms prior methods in terms of final performance and stability, particularly for the first four environments where V-MAIL reaches near-expert performance. In the most challenging visual Baoding Balls task, which is notably difficult even with ground-truth state, only V-MAIL is able to make some progress, but all methods struggle. Confidence intervals are shown with 1 SE over 6 runs.}
    \label{fig:results}
    \vspace{-0.3cm}
\end{figure}

Experiment results are shown in Figure \ref{fig:results}. To answer questions (1) and (2), we compare V-MAIL to DA-SQIL and DA-DAC on the Cheetah and Walker tasks. We find that V-MAIL efficiently and reliably solves both tasks; in contrast, the model-free methods initially outperform V-MAIL, but their performance has high variance across random seeds and exhibits significant instability. Such stability issues have also been observed by \citet{swamy2021moments}, which provides some theoretical explanation in the case of SQIL and the suggestion of early stopping as a mitigation technique. In the case of DAC, the reasons for instability are less clear. %While performance collapse of SQIL is somewhat regular, DAC shows high variance between random seeds.
%%CF: I'm not sure I agree with this. The error bars for SQIL are actually usually larger than those for DAC
%Collapse in performance seems to be caused by divergence of the critic, rather than of the discriminator.
%%CF: If you are going to say this, then you need to point to something from which you are drawing this conclusion (e.g. putting a plot of the critic loss). For now, I commented it out.
%This would suggest that the problem originates from some complex interaction between the non-stationary reward and bootstrapping error of Q-learning. 
Motivated by instability we observed in the critic loss for DA-DAC, we experimented with a number of mitigation strategies in an attempt to improve DA-DAC, including constraining the discriminator, varying the buffer and batch sizes, and separating the convolutional encoders of the discriminator and the actor/critic; however, these techniques didn't fully prevented the degradation in performance.

%%CF: Should answer questions in the order that they are introduced. Also, this answer to question (5) is way to informal. Should at least point the reader to an appendix with a plot or table. Otherwise, I wouldn't include it.
%%RR: It's a bit hard to see but each of those is a different curve in the first two grphs in Figure 3.

On the Car Racing environment, we find that DA-SQIL and DA-DAC can reach or outperform behavior cloning, but struggle to reach expert-level performance. In contrast, V-MAIL stably and reliably achieves near-expert performance in about 200k environment steps. Note that \citet{SQIL2020Reddy} report expert-level performance on this task, but in an easier setting with double the number of expert demonstrations available (20 vs. 10). Given that tracks are randomly generated per episode demanding significant generalization, it is not surprising that the problem becomes considerably more difficult with only 10 demonstrations.
% on this task, however we haven't been able to reproduce those results. Unlike the original experiments we use only 10 trajectories, rather than 20, also our expert is of much higher quality reaching average performance of 922.3, versus 704 for the data used in SQIL. This suggest that SQIL needs a wider distribution of data to solve the task. This seems to be especially relevant in the Car Racing domain as tracks are randomly generated and the agent needs to learn to race across random configurations, which it may not have seen during training. 

Finally, to answer question (3), we consider the D'Claw and Baoding Balls tasks. In the D'Claw environment, SQIL fails to make progress, while DA-DAC makes significant progress initially but quickly degrades. V-MAIL solves the task in less than 100k environment steps. In the most challenging visual Baoding Balls problem, involving a 26-dimensional control space, V-MAIL is the only algorithm to reach any success.
%Finally the Baoding Balls is an extremely challenging task with 26-dimensional control space and needs to learn a contact model between the palm, fingers, each individual ball and between the balls themselves. V-MAIL is the only algorithm to reach any success on this domain. 

\subsection{Ablation Experiments}
\begin{figure}
    \centering
    \includegraphics[height=5.0cm]{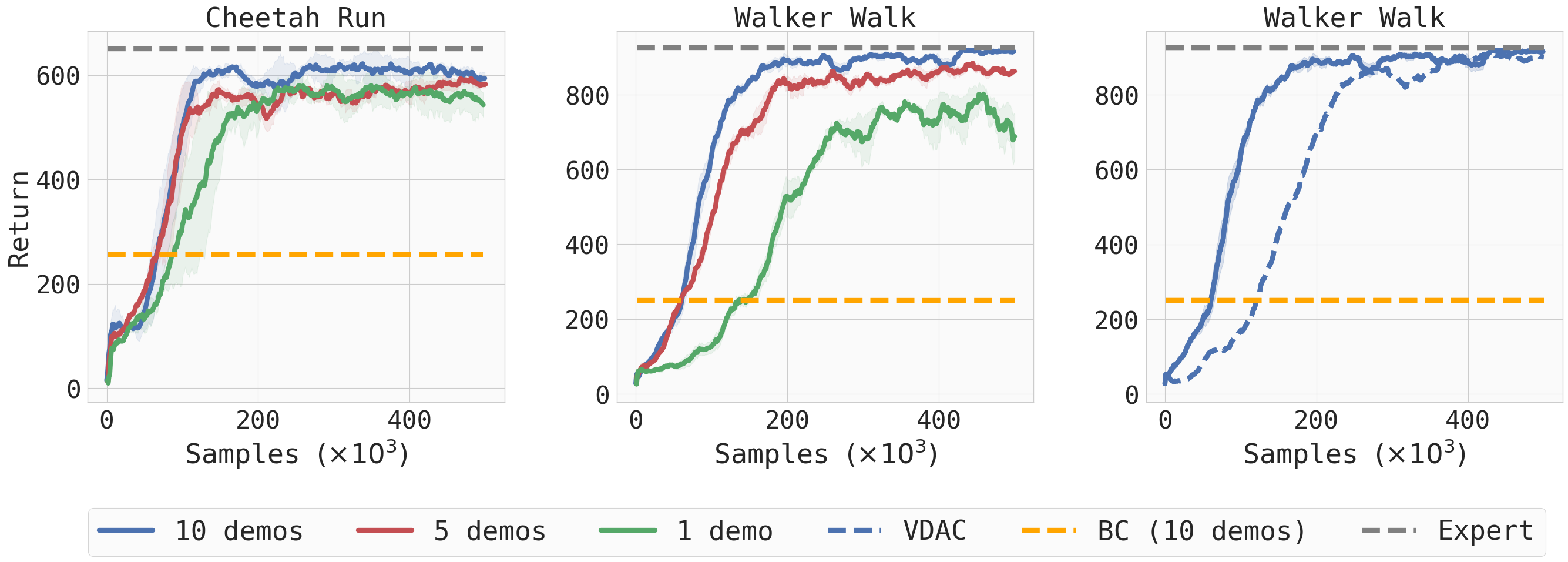}
    \caption{Ablation experiments for the VMAIL model. The left and middle graph show the efficacy of learning from different number of demonstrations where VMAIL outperforms baseline models even with a single demo. The final figure compares VMAIL to a model which used the variatonal model for representation purposes only (labeled VDAC). We see that VMAIL achieves 30\% higher sample efficiency. Confidence intervals are shown with 1 SE over 3 runs.}
    \label{fig:ablations}
\end{figure}

Results for our ablation experiments are shown in Fig. \ref{fig:ablations}.

\textbf{Number of Demonstrations.} In the first set of experiments we evaluate VMAIL's capability to learn from a limited number of demonstrations. We deploy our model on the two locomotion environments using 10, 5 and a single demonstration trajectory. VMAIL shows only minor deterioration in performance on the Cheetah Run environment, as well as on the Walker Walk environment when using only 5 demos, but it struggles to reach expert-level performance on the walker task when provided with a single trajectory. However, in all cases VMAIL outperforms the asymptotic results of our baselines, which use the full 10 demonstrations.

\textbf{Effect of Representation Learning.} We also evaluate the effect of representation learning on model performance. In this scenario we still train a variational model, but similar to \cite{SLAC20202Lee} only use it for for representation purposes. We then train a standard Discriminator-Actor Critic model on top of the learned latent space, which we denote as Variational DAC or VDAC. Results are shown in the last graph of Fig. \ref{fig:ablations}. VDAC exhibits more stable performance as compared to learning directly from pixels, however it is still 30\% less sample efficient than VMAIL.

\subsection{Transfer Experiments}
\label{sec:transfer_experiments}

\iffalse
\begin{table}[]
    \centering
    \begin{tabular}{ p{2.0cm}|p{1.5cm}|p{1.5cm}|p{1.5cm}|p{1.5cm}|p{1.5cm}  }
    \toprule
    Domain & Target Task IL & VMAIL (ours) & Offline DAC & Behaviour cloning & Policy Transfer\\
    \midrule
    Walker Walk  & 98.2\%    & \textbf{92.7\%} &  8.8\% & 26.8\%  &21.3\%\\
    Claw Rotate  & 102.3\%   & \textbf{97.9\%}&  -0.7\%& 8.3\%   &5.6\% \\
    \bottomrule
    \end{tabular}
    \vspace{1mm}
    %%CF.5.24: I would recommend transposing this table, and having it be a wrap table (or using a bar plot)
    %%CF.5.24: is target task IL an oracle? if so, you should mark it as such and provide separation between it and the other methods.
    \caption{\small Performance on zero-shot transfer to a new imitation learning task as percent of expert return. Each method is provided with 10 demonstrations of the target task, and zero additional samples in the environment. V-MAIL can solve the target tasks within its learned model without any additional samples, while approaches such as offline model-free learning using source samples, behavior cloning on the target task demonstrations, and direct policy transfer from the source tasks do not effectively solve the tasks. }
    \label{tab:zeroshot}
\end{table}
\fi

%

%%CF: TODO - should make this a wrap table probably

\textbf{Transfer Scenarios.} To evaluate V-MAIL's ability to learn new imitation tasks in a zero-shot way (i.e. without any additional environment samples) we deploy Algorithm \ref{alg:fewshot} on two domains: in a locomotion experiment we train on the Walker Stand and Walker Run (target speed greater than 8) tasks and and evaluate transfer to the Walker Walk (target speed between 2 and 4) task from the DeepMind Control suite. In a manipulation scenario, we use a set of custom D'Claw Screw tasks from the Robel suite \citep{Kumar_ROBEL}. We train our model on the 3-prong tasks with clockwise and counter-clockwise rotation, as well as the 4-prong task with counter-clockwise rotation and evaluate transfer to the 4-prong task with clockwise rotation. 

\textbf{Comparisons.} We devise several points of comparison. First, we compare to directly applying the policy learned in the most related source task to the target task. This tests whether the target task demands qualitatively distinct behavior. Second, we compare to an offline version of DAC, augmented with the CQL approach \cite{kumar2020conservative}, where samples collected from the source task are used to update the policy, with the target task demonstrations used to learn the reward. Finally, we also compare to behavior cloning on the target task demonstrations (without leveraging any source task data), and an oracle that performs V-MAIL on the target task directly.

\begin{wraptable}{r}{7.5cm}
\small
\vspace{-0.3cm}
%\begin{table}[]
    \centering
    \begin{tabular}{ l| c | c }
    \toprule
    Method & Walker  Walk & Claw Rotate  \\
    %Method & Walker \{Stand, Run\} $\rightarrow$ Walk & Claw Rotate \{3-cw, 3-ccw, 4-ccw\} $\rightarrow$ 4-cw \\
    \midrule
    Offline DAC  &     8.8\% &  -0.7\%  \\
    Behavior cloning & 26.8\% &  8.3\%  \\
    Policy transfer & 21.3\% & 5.6\% \\
    V-MAIL (ours)  & \textbf{92.7\%} &  \textbf{97.9\%}     \\
    \midrule
    Target task IL (oracle) & 98.2\% & 102.3\% \\
    \bottomrule
    \end{tabular}
    %\vspace{1mm}
    \caption{\small Zero-shot transfer performance to a new imitation learning task as percent of expert return. Each method is provided with 10 demonstrations of the target task, and zero additional environment samples. V-MAIL can solve the target tasks within its learned model without any additional samples, while model-free transfer learning approaches fail.}
    \vspace{-0.2cm}
    \label{tab:zeroshot}
%\end{table}
\end{wraptable}
\textbf{Results.} The results in Table~\ref{tab:zeroshot}. Policy transfer performs poorly, suggesting that the target task indeed requires qualitatively different behaviour from the few training tasks available. Further, behavior cloning on the target demonstrations is not sufficient to learn the task. Offline DAC also shows poor performance.
Finally, we see that V-MAIL almost matches the performance of the agent explicitly trained on task, indicating the learned model and the algorithm for training within that model can be used not just for efficient visual imitation learning, but also for zero-shot transfer to new tasks.

\section{Related Work}
% Here, we review the relevant literature on imitation learning and image-based RL.
\paragraph{Imitation Learning.}
%%CF.5.7: It seems like this paragraph only covers adversarial imitation learning. It would be good to cover non-adversarial imitation learning approaches too, like BC and inverse RL, including methods that operate on image inputs (e.g. ALVINN, the NVIDIA driving paper)
%%CF.5.7: There are also some papers that do GAIL with image observations. It would be good to try to track those down.
%%CF.5.7: There's also a model-based GAIL paper that appeared in ICML 2017 I think. This one: http://proceedings.mlr.press/v70/baram17a.html
%%CF.5.7: I would recommend doing a little literature search to find relevant papers. For example, this paper also looks potentially relevant, which I found through a google scholar search: https://ieeexplore.ieee.org/abstract/document/9361118/

Recent model-free imitation learning can be categorized as either adversarial or non-adversarial. Adversarial methods inspired by GANs \cite{Goodfellow2014GenerativeAN} train an explicit classifier between expert and policy behaviour and optimize the agent in a two-player minimax game. GAIL \cite{GAIL2016Ho} and AIRL \cite{AIRLFu2018} are two such algorithms; however they often have poor sample efficiency due to the requirement of on-policy rollouts in the environment. To address sample efficiency issues, off-policy variants such as DAC \cite{DAC2019Kostrikov} and SAM \cite{SAM2019Blonde} have been developed, however they suffer from an objective mismatch when using off-policy data~\citep{ValueDICE2019Kostrikov}, often resulting in learning instability~\cite{blonde2020lipschitzness}. 
% A common issue in all adversarial methods is the difficulty and (and instability) of optimizing the minimax objective.

An alternate line of research attempts to forego adversarial training: SQIL \cite{SQIL2020Reddy} frames the problem as regularized behaviour cloning and trains an off-policy algorithm with rewards of 1 for expert trajectories and 0 for policy ones. RCE \cite{eysenbach2021replacing} uses a very similar approach, but derives it through a different objective. ValueDICE \cite{ValueDICE2019Kostrikov} uses the same key result for iterative distribution matching as RCE to obtain an off-policy distribution matching algorithm. In \citet{swamy2021moments} the authors derive distribution matching as a bound on policy under-performance, similar to our analysis in Section \ref{sec:algo_mdp} and propose a practical non-adversarial algorithm AdVIL, however in reported experiments it does not outperform behaviour cloning. In \citet{berseth2021learning} the authors consider a variational encoder, which embeds expert video demonstrations in a latent space, and optimize a mis-match objective using on-policy model-free training. 

A few papers have considered model-based imitation learning as well: \citet{baram2016modelbased} is an adversarial algorithm conceptually similar to our approach, but only focuses on low-dimensional state-based tasks and train the discriminator using off-policy replay buffer, which does not allow it to generalize to new tasks. Related to our method is \citet{finn2016guided} which uses a similar reward learning in combination with a locally linear dynamics model, which leads to trajectory-centric algorithms that cannot transfer the model to new tasks. \citet{das2021modelbased} considers a similar setting for inverse RL using a simplified parameterization of the cost function. In this work we develop end-to-end model for adversarial imitation learning in high-dimensional POMDPs and generalization to novel tasks without hand-designed features. 
%%CF.5.7: Seems a little bit weird to cover finn2016 last given that it comes before all of these works.
A related line of research is learning from observations alone (without access to expert actions). The BCO approach \citet{torabi2018behavioral} learns an inverse dynamics model to infer and copy expert actions, however it still suffers from the drawbacks of behaviour cloning. In a concurrent work \cite{kidambi2021mobile} explore similar objective to \cite{chang2021mitigating} and VMAIL, using only states for training the discriminator objective, however they only focus on low dimensional domains. With a particular variational model specification VMAIL can be also be extended to work only with expert state observations, which we leave for further work.   

Finally, related to zero-shot imitation learning, concurrent work by \citet{chang2021mitigating} extends a model-based offline RL algorithm~\citep{Kidambi-MOReL-20, MOPO} to imitation learning. The focus in their work is primarily theoretical analysis, with empirical results in tasks with compact state representations. In contrast, our work aims to develop a stable and efficient imitation learning algorithm that can handle high-dimensional observation spaces (like visual inputs), in both online and offline learning settings. 
%To our knowledge, no prior work has considered this zero-shot IL approach in high-dimensional settings.

\noindent \textbf{Reinforcement Learning From Images with Variational Models.}
%%CF.5.7: maybe change this to model-based reinforcement learning from images, or variational models for reinforcement learning? I think it would be good to be more specific in some way. 
%%CF.5.7: BEE, plan2explore, and LOMPO are potentially relevant here too?
Reinforcement learning from images is an inherently difficult task, since the agent needs to learn meaningful visual representations to support policy learning. A recent line of research \citep{gelada2019deepmdp, PlanNet2019Hafner, SLAC20202Lee, Dreamer2020Hafner,rafailov2020offline} train a variational model of the image-based environment as an auxiliary task, either for representation learning only~\cite{gelada2019deepmdp, SLAC20202Lee} or for additionally generating on-policy data by rolling out the model~\cite{Dreamer2020Hafner}. Our method builds upon these ideas, but unlike these prior works, considers the problem of learning from visual demonstrations without access to rewards.
%%CF.5.7: I don't think you need to be this verbose. Could just add a phrase to the end of the 2nd sentence saying "either for representation learning only~\cite{deepmdp,slac} or for generating on-policy data by rolling out the model~\cite{hafner}."
%%CF.5.7: I don't think the reader will understand what you mean by the correct objective without more context
%\citet{kostrikov2020image} introduce a sample efficient model-free approach for off-policy Q-learning from images using data augmentation. In our experiments, we combine this approach with model-free imitation learning methods in order to strengthen our points of comparison. 
%%CF: I think we probably don't need this last part, so I'm commenting it out.
%The authors report state of the art performance on both asymptotic returns and improve sample efficiency by up to two orders of magnitude. 

%%CF.5.7: This is way more info on DrQ than you need. If you think it's important to mention SOTA, than you can add a phrase to the first sentence "achieving state-of-the-art returns and sample efficiency"
%%CF.5.7: should just give a section number.

\section{Conclusion}
In this work we presented V-MAIL, a model-based imitation learning algorithm that works from high-dimensional image observations. V-MAIL learns a model of the environment, which serves as a strong self-supervision signal for visual representation learning and mitigates distribution shift by enabling synthetic on-policy  rollouts using the model. Through experiments, we find that V-MAIL achieves better asymptotic performance, is more stable, and matches the sample efficiency of prior model-free approaches. By effectively re-using the learned model, V-MAIL is also successful in zero-shot imitation learning, capable of learning new tasks using a small number of demonstrations, without any additional interactions with the environment.
% by training a policy using only synthetic model rollouts, our approach is a strong procedure for zero-shot transfer to novel imitation learning tasks.

% . One potential approach is to combine offline reinforcement learning approaches like \citet{rafailov2020offline}, to scale offline imitation learning to high-dimensional domains. Finally, our experiments suggest that this algorithm is efficient enough to be applied to real robots, an interesting direction for future work.
%We expect that the generalisation capabilities of our approach will be applicable in multi-task, continuous imitation and inverse meta-reinforcement learning as well.
%%CF: I don't think this last sentence is really saying much

\textbf{Limitations.} VMAIL uses a low-dimensional dynamics model, explicitly trained with image prediction. This might make our method vulnerable to adversarial visual perturbations. Although successful in domains with complex dynamics, our approach relies on variational models with compact, single-level, latent state spaces. It is possible that such a model class may not have sufficient capacity to represent complex scenes with multiple cluttered objects or deformable objects. A different selection of training objectives or further developments and improvements in variational generative modeling can potentially address both of these limitations.

\textbf{Future Work.} 
We believe this work opens the door for multiple potential developments. One direction is to train our procedure using only expert observations without access to expert actions, which is an even more realistic scenario. This setup is quite difficult for model-free approaches, since expert actions usually serve as a strong supervision. Another direction is to use on-policy model based rollouts to efficiently train other algorithms that inherently require on-policy data, such as multi-modal imitation~\citep{li2017infogail, hausman2017multimodal}. We showed the transfer capabilities of our algorithm to new tasks in a zero-shot imitation learning formulation. However, V-MAIL can in principle utilize any previously collected data for model-training, enabling potential applications in offline imitation learning in conjunction with offline RL algorithms like \citet{rafailov2020offline}.

\section*{Acknowledgements}
%%AR: Note made some edits to Acknowledgements since FB legal wanted this wording.
This work was supported in part byt  ONR grants N00014-20-1-2675 and  N00014-21-1-2685 as well as Intel Corporation. Chelsea Finn is a CIFAR Fellow in the Learning in Machines and Brains program. Part of this work was completed when Aravind Rajeswaran was at the University of Washington, where he was supported through the J.P.~Morgan PhD Fellowship in AI (2020-21).

\bibliography{neurips_2021}

\begin{thebibliography}{57}
\providecommand{\natexlab}[1]{#1}
\providecommand{\url}[1]{\texttt{#1}}
\expandafter\ifx\csname urlstyle\endcsname\relax
  \providecommand{\doi}[1]{doi: #1}\else
  \providecommand{\doi}{doi: \begingroup \urlstyle{rm}\Url}\fi

\bibitem[Amodei et~al.(2016)Amodei, Olah, Steinhardt, Christiano, Schulman, and
  Man{\'e}]{Amodei2016ConcretePI}
Dario Amodei, Chris Olah, J.~Steinhardt, Paul~F. Christiano, John Schulman, and
  Dan Man{\'e}.
\newblock Concrete problems in ai safety.
\newblock \emph{ArXiv}, abs/1606.06565, 2016.

\bibitem[Everitt and Hutter(2019)]{Everitt2019RewardTP}
Tom Everitt and Marcus Hutter.
\newblock Reward tampering problems and solutions in reinforcement learning: A
  causal influence diagram perspective.
\newblock \emph{ArXiv}, abs/1908.04734, 2019.

\bibitem[Rajeswaran et~al.(2018)Rajeswaran, Kumar, Gupta, Vezzani, Schulman,
  Todorov, and Levine]{Rajeswaran-RSS-18}
Aravind Rajeswaran, Vikash Kumar, Abhishek Gupta, Giulia Vezzani, John
  Schulman, Emanuel Todorov, and Sergey Levine.
\newblock {Learning Complex Dexterous Manipulation with Deep Reinforcement
  Learning and Demonstrations}.
\newblock In \emph{Proceedings of Robotics: Science and Systems (RSS)}, 2018.

\bibitem[Portelas et~al.(2020)Portelas, Colas, Weng, Hofmann, and
  Oudeyer]{Portelas2020AutomaticCL}
R{\'e}my Portelas, C{\'e}dric Colas, Lilian Weng, Katja Hofmann, and
  Pierre-Yves Oudeyer.
\newblock Automatic curriculum learning for deep rl: A short survey.
\newblock \emph{ArXiv}, abs/2003.04664, 2020.

\bibitem[Khetarpal et~al.(2020)Khetarpal, Riemer, Rish, and
  Precup]{Khetarpal2020TowardsCR}
Khimya Khetarpal, Matthew Riemer, I.~Rish, and Doina Precup.
\newblock Towards continual reinforcement learning: A review and perspectives.
\newblock \emph{ArXiv}, abs/2012.13490, 2020.

\bibitem[Lowe et~al.(2017)Lowe, Wu, Tamar, Harb, Abbeel, and
  Mordatch]{Lowe2017MultiAgentAF}
Ryan Lowe, Yi~Wu, Aviv Tamar, Jean Harb, P.~Abbeel, and Igor Mordatch.
\newblock Multi-agent actor-critic for mixed cooperative-competitive
  environments.
\newblock In \emph{NIPS}, 2017.

\bibitem[Pomerleau(1988)]{pomerleau1988alvinn}
Dean~A Pomerleau.
\newblock Alvinn: an autonomous land vehicle in a neural network.
\newblock In \emph{Proceedings of the 1st International Conference on Neural
  Information Processing Systems}, pages 305--313, 1988.

\bibitem[Ross et~al.(2011)Ross, Gordon, and Bagnell]{Dagger2011Ross}
Stephane Ross, Geoffrey~J. Gordon, and J.~Andrew Bagnell.
\newblock A reduction of imitation learning and structured prediction to
  no-regret online learning.
\newblock \emph{AISTATS}, 2011.

\bibitem[Spencer et~al.(2021)Spencer, Choudhury, Venkatraman, Ziebart, and
  Bagnell]{Feedback2021Spencer}
Jonathan Spencer, Sanjiban Choudhury, Arun Venkatraman, Brian Ziebart, and
  J.~Andrew Bagnell.
\newblock Feedback in imitation learning: The three regimes of covariate shift.
\newblock \emph{ArXiv Preprint}, 2021.

\bibitem[Finn et~al.(2016{\natexlab{a}})Finn, Levine, and
  Abbeel]{finn2016guided}
Chelsea Finn, Sergey Levine, and Pieter Abbeel.
\newblock Guided cost learning: Deep inverse optimal control via policy
  optimization.
\newblock In \emph{International conference on machine learning}, pages 49--58.
  PMLR, 2016{\natexlab{a}}.

\bibitem[Fu et~al.(2018)Fu, Luo, and Levine]{AIRLFu2018}
Justin Fu, Katie Luo, and Sergey Levine.
\newblock Learning robust rewards with adversarial inverse reinforcement
  learning.
\newblock \emph{International Conference on Learning Representations}, 2018.

\bibitem[Ho and Ermon(2016)]{GAIL2016Ho}
Jonathan Ho and Stefano Ermon.
\newblock Generative adversarial imitation learning.
\newblock \emph{Conference on Neural Information Processing Systems}, 2016.

\bibitem[Finn et~al.(2016{\natexlab{b}})Finn, Christiano, Abbeel, and
  Levine]{GAIL201Finn}
Chelsea Finn, Paul Christiano, Pieter Abbeel, and Sergey Levine.
\newblock A connection between generative adversarial networks, inverse
  reinforcement learning, and energy-based models.
\newblock \emph{ArXiv Preprint}, 2016{\natexlab{b}}.

\bibitem[Ghasemipour et~al.(2019)Ghasemipour, Zemel, and Gu]{DIV2019Sayed}
Seyed Kamyar~Seyed Ghasemipour, Richard Zemel, and Shixiang Gu.
\newblock A divergence minimization perspective on imitation learning methods.
\newblock \emph{Conference on Robot Learning}, 2019.

\bibitem[Goodfellow et~al.(2014)Goodfellow, Pouget-Abadie, Mirza, Xu,
  Warde-Farley, Ozair, Courville, and Bengio]{Goodfellow2014GenerativeAN}
Ian~J. Goodfellow, Jean Pouget-Abadie, M.~Mirza, Bing Xu, David Warde-Farley,
  Sherjil Ozair, Aaron~C. Courville, and Yoshua Bengio.
\newblock Generative adversarial nets.
\newblock In \emph{NIPS}, 2014.

\bibitem[Ke et~al.(2019)Ke, Barnes, Sun, Lee, Choudhury, and
  Srinivasa]{Ke2019ImitationLA}
Liyiming Ke, Matt Barnes, W.~Sun, Gilwoo Lee, Sanjiban Choudhury, and
  S.~Srinivasa.
\newblock Imitation learning as f-divergence minimization.
\newblock \emph{ArXiv}, abs/1905.12888, 2019.

\bibitem[Ziebart et~al.(2008)Ziebart, Maas, Bagnell, and
  Dey]{Ziebart2008MaximumEI}
Brian~D. Ziebart, Andrew~L. Maas, J.~Bagnell, and A.~Dey.
\newblock Maximum entropy inverse reinforcement learning.
\newblock In \emph{AAAI}, 2008.

\bibitem[Kostrikov et~al.(2019)Kostrikov, Agrawal, Dwibedi, Levine, and
  Tompson]{DAC2019Kostrikov}
Ilya Kostrikov, Kumar~Krishna Agrawal, Debidatta Dwibedi, Sergey Levine, and
  Jonathan Tompson.
\newblock Discriminator-actor-critic: Addressing sample inefficiency and reward
  bias in adversarial imitation learning.
\newblock \emph{International Conference on Learning Representations}, 2019.

\bibitem[Blondé and Kalousis(2019)]{SAM2019Blonde}
Lionel Blondé and Alexandros Kalousis.
\newblock Sample-efficient imitation learning via generative adversarial nets.
\newblock \emph{AISTATS}, 2019.

\bibitem[Brock et~al.(2019)Brock, Donahue, and Simonyan]{Brock2019LargeSG}
Andrew Brock, Jeff Donahue, and K.~Simonyan.
\newblock Large scale gan training for high fidelity natural image synthesis.
\newblock \emph{ArXiv}, abs/1809.11096, 2019.

\bibitem[Rajeswaran et~al.(2020)Rajeswaran, Mordatch, and
  Kumar]{RajeswaranGameMBRL}
Aravind Rajeswaran, Igor Mordatch, and Vikash Kumar.
\newblock {A Game Theoretic Framework for Model-Based Reinforcement Learning}.
\newblock In \emph{{ICML}}, 2020.

\bibitem[Gangwani et~al.(2020)Gangwani, Lehman, Liu, and
  Peng]{gangwani2019learning}
Tanmay Gangwani, Joel Lehman, Qiang Liu, and Jian Peng.
\newblock Learning belief representations for imitation learning in pomdps.
\newblock \emph{Conference on Uncertainty in Artificial Intelligence}, 2020.

\bibitem[Watter et~al.(2015)Watter, Springenberg, Boedecker, and
  Riedmiller]{watter2015embed}
Manuel Watter, Jost~Tobias Springenberg, Joschka Boedecker, and Martin
  Riedmiller.
\newblock Embed to control: A locally linear latent dynamics model for control
  from raw images.
\newblock \emph{Conference on Neural Information Processing Systems}, 2015.

\bibitem[Zhang et~al.(2019)Zhang, Vikram, Smith, Abbeel, Johnson, and
  Levine]{zhang2019solar}
Marvin Zhang, Sharad Vikram, Laura Smith, Pieter Abbeel, Matthew~J. Johnson,
  and Sergey Levine.
\newblock Solar: Deep structured representations for model-based reinforcement
  learning.
\newblock \emph{International Conference on Machine Learning}, 2019.

\bibitem[Lee et~al.(2020)Lee, Nagabandi, Abbeel, and Levine]{SLAC20202Lee}
Alex~X. Lee, Anusha Nagabandi, Pieter Abbeel, and Sergey Levine.
\newblock Stochastic latent actor-critic: Deep reinforcement learning with a
  latent variable model.
\newblock \emph{Conference on Neural Information Processing Systems}, 2020.

\bibitem[Gelada et~al.(2019)Gelada, Kumar, Buckman, Nachum, and
  Bellemare]{gelada2019deepmdp}
Carles Gelada, Saurabh Kumar, Jacob Buckman, Ofir Nachum, and Marc~G.
  Bellemare.
\newblock Deepmdp: Learning continuous latent space models for representation
  learning.
\newblock \emph{International Conference on Machine Learning}, 2019.

\bibitem[Hafner et~al.(2019)Hafner, Lillicrap, Fischer, Villegas, Ha, Lee, and
  Davidson]{PlanNet2019Hafner}
Danijar Hafner, Timothy Lillicrap, Ian Fischer, Ruben Villegas, David Ha,
  Honglak Lee, and James Davidson.
\newblock Learning latent dynamics for planning from pixels.
\newblock \emph{International Conference on Machine Learning}, 2019.

\bibitem[Hafner et~al.(2020)Hafner, Lillicrap, Ba, and
  Norouzi]{Dreamer2020Hafner}
Danijar Hafner, Timothy Lillicrap, Jimmy Ba, and Mohammad Norouzi.
\newblock Dream to control: Learning behaviors by latent imagination.
\newblock \emph{International Conference on Learning Representations}, 2020.

\bibitem[Reddy et~al.(2020)Reddy, Dragan, and Levine]{SQIL2020Reddy}
Siddharth Reddy, Anca~D. Dragan, and Sergey Levine.
\newblock Sqil: Imitation learning via reinforcement learning with sparse
  rewards.
\newblock \emph{International Conference on Learning Representations}, 2020.

\bibitem[Tassa et~al.(2018)Tassa, Doron, Muldal, Erez, Li, de~Las~Casas,
  Budden, Abdolmaleki, Merel, Lefrancq, Lillicrap, and
  Riedmiller]{tassa2018deepmind}
Yuval Tassa, Yotam Doron, Alistair Muldal, Tom Erez, Yazhe Li, Diego
  de~Las~Casas, David Budden, Abbas Abdolmaleki, Josh Merel, Andrew Lefrancq,
  Timothy Lillicrap, and Martin Riedmiller.
\newblock Deepmind control suite, 2018.

\bibitem[Brockman et~al.(2016)Brockman, Cheung, Pettersson, Schneider,
  Schulman, Tang, and Zaremba]{brockman2016openai}
Greg Brockman, Vicki Cheung, Ludwig Pettersson, Jonas Schneider, John Schulman,
  Jie Tang, and Wojciech Zaremba.
\newblock Openai gym, 2016.

\bibitem[Ahn et~al.(2019)Ahn, Zhu, Hartikainen, Ponte, Gupta, Levine, and
  Kumar]{Kumar_ROBEL}
Michael Ahn, Henry Zhu, Kristian Hartikainen, Hugo Ponte, Abhishek Gupta,
  Sergey Levine, and Vikash Kumar.
\newblock {ROBEL: RObotics BEnchmarks for Learning with low-cost robots}.
\newblock In \emph{Conference on Robot Learning (CoRL)}, 2019.

\bibitem[Swamy et~al.(2021)Swamy, Choudhury, Wu, and Bagnell]{swamy2021moments}
Gokul Swamy, Sanjiban Choudhury, Zhiwei~Steven Wu, and J.~Andrew Bagnell.
\newblock Of moments and matching: Trade-offs and treatments in imitation
  learning.
\newblock \emph{International Conference on Machine Learning}, 2021.

\bibitem[Kumar et~al.(2020)Kumar, Zhou, Tucker, and
  Levine]{kumar2020conservative}
Aviral Kumar, Aurick Zhou, George Tucker, and Sergey Levine.
\newblock Conservative q-learning for offline reinforcement learning.
\newblock \emph{Conference on Neural Information Processing Systems}, 2020.

\bibitem[Kostrikov et~al.(2020{\natexlab{a}})Kostrikov, Nachum, and
  Tompson]{ValueDICE2019Kostrikov}
Ilya Kostrikov, Ofir Nachum, and Jonathan Tompson.
\newblock Imitation learning via off-policy distribution matching.
\newblock \emph{International Conference on Learning Representations},
  2020{\natexlab{a}}.

\bibitem[Blondé et~al.(2020)Blondé, Strasser, and
  Kalousis]{blonde2020lipschitzness}
Lionel Blondé, Pablo Strasser, and Alexandros Kalousis.
\newblock Lipschitzness is all you need to tame off-policy generative
  adversarial imitation learning, 2020.

\bibitem[Eysenbach et~al.(2021)Eysenbach, Levine, and
  Salakhutdinov]{eysenbach2021replacing}
Benjamin Eysenbach, Sergey Levine, and Ruslan Salakhutdinov.
\newblock Replacing rewards with examples: Example-based policy search via
  recursive classification, 2021.

\bibitem[Berseth et~al.(2021)Berseth, Golemo, and Pal]{berseth2021learning}
Glen Berseth, Florian Golemo, and Christopher Pal.
\newblock Towards learning to imitate from a single video demonstration, 2021.

\bibitem[Baram et~al.(2016)Baram, Anschel, and Mannor]{baram2016modelbased}
Nir Baram, Oron Anschel, and Shie Mannor.
\newblock Model-based adversarial imitation learning.
\newblock \emph{Conference on Neural Information Processing Systems}, 2016.

\bibitem[Das et~al.(2020)Das, Bechtle, Davchev, Jayaraman, Rai, and
  Meier]{das2021modelbased}
Neha Das, Sarah Bechtle, Todor Davchev, Dinesh Jayaraman, Akshara Rai, and
  Franziska Meier.
\newblock Model-based inverse reinforcement learning from visual
  demonstrations.
\newblock \emph{Conference on Robot Learning}, 2020.

\bibitem[Torabi et~al.(2018)Torabi, Warnell, and Stone]{torabi2018behavioral}
Faraz Torabi, Garrett Warnell, and Peter Stone.
\newblock Behavioral cloning from observation.
\newblock \emph{International Joint Conference on Artificial Intelligence},
  2018.

\bibitem[Kidambi et~al.(2021)Kidambi, Chang, and Sun]{kidambi2021mobile}
Rahul Kidambi, Jonathan Chang, and Wen Sun.
\newblock Mobile: Model-based imitation learning from observation alone, 2021.

\bibitem[Chang et~al.(2021)Chang, Uehara, Sreenivas, Kidambi, and
  Sun]{chang2021mitigating}
Jonathan~D. Chang, Masatoshi Uehara, Dhruv Sreenivas, Rahul Kidambi, and Wen
  Sun.
\newblock Mitigating covariate shift in imitation learning via offline data
  without great coverage, 2021.

\bibitem[Kidambi et~al.(2020)Kidambi, Rajeswaran, Netrapalli, and
  Joachims]{Kidambi-MOReL-20}
Rahul Kidambi, Aravind Rajeswaran, Praneeth Netrapalli, and Thorsten Joachims.
\newblock Morel : Model-based offline reinforcement learning.
\newblock In \emph{NeurIPS}, 2020.

\bibitem[Yu et~al.(2020)Yu, Thomas, Yu, Ermon, Zou, Levine, Finn, and Ma]{MOPO}
Tianhe Yu, Garrett Thomas, Lantao Yu, Stefano Ermon, James Zou, Sergey Levine,
  Chelsea Finn, and Tengyu Ma.
\newblock Mopo: Model-based offline policy optimization.
\newblock \emph{CoRR}, abs/2005.13239, 2020.

\bibitem[Rafailov et~al.(2020)Rafailov, Yu, Rajeswaran, and
  Finn]{rafailov2020offline}
Rafael Rafailov, Tianhe Yu, Aravind Rajeswaran, and Chelsea Finn.
\newblock Offline reinforcement learning from images with latent space models.
\newblock \emph{arXiv preprint arXiv:2012.11547}, 2020.

\bibitem[Li et~al.(2017)Li, Song, and Ermon]{li2017infogail}
Yunzhu Li, Jiaming Song, and Stefano Ermon.
\newblock Infogail: Interpretable imitation learning from visual
  demonstrations.
\newblock \emph{Conference on Neural Information Processing Systems}, 2017.

\bibitem[Hausman et~al.(2017)Hausman, Chebotar, Schaal, Sukhatme, and
  Lim]{hausman2017multimodal}
Karol Hausman, Yevgen Chebotar, Stefan Schaal, Gaurav Sukhatme, and Joseph Lim.
\newblock Multi-modal imitation learning from unstructured demonstrations using
  generative adversarial nets, 2017.

\bibitem[Ali and Silvey(1966)]{ali166datainequality}
Syed~Mumtaz Ali and Samuel Silvey.
\newblock . a general class of coefficients of divergence of one distribution
  from another.
\newblock \emph{Journal of the Royal Statistical Society}, 28(1):\penalty0
  131:142, 1966.

\bibitem[Karl et~al.(2017)Karl, Soelch, Bayer, and van~der Smagt]{karl2017deep}
Maximilian Karl, Maximilian Soelch, Justin Bayer, and Patrick van~der Smagt.
\newblock Deep variational bayes filters: Unsupervised learning of state space
  models from raw data.
\newblock \emph{International Conference on Machine Learning}, 2017.

\bibitem[Blei et~al.(2016)Blei, Kucukelbir, and
  McAuliffe]{Blei2016VariationalIA}
David~M. Blei, A.~Kucukelbir, and Jon~D. McAuliffe.
\newblock Variational inference: A review for statisticians.
\newblock \emph{Journal of the American Statistical Association}, 112:\penalty0
  859 -- 877, 2016.

\bibitem[Kingma and Welling(2014)]{kingma2014autoencoding}
Diederik~P Kingma and Max Welling.
\newblock Auto-encoding variational bayes, 2014.

\bibitem[Feinberg et~al.(2018)Feinberg, Wan, Stoica, Jordan, Gonzalez, and
  Levine]{feinberg2018modelbased}
Vladimir Feinberg, Alvin Wan, Ion Stoica, Michael~I. Jordan, Joseph~E.
  Gonzalez, and Sergey Levine.
\newblock Model-based value estimation for efficient model-free reinforcement
  learning.
\newblock \emph{International Conference on Machine Learning}, 2018.

\bibitem[Buckman et~al.(2019)Buckman, Hafner, Tucker, Brevdo, and
  Lee]{buckman2019sampleefficient}
Jacob Buckman, Danijar Hafner, George Tucker, Eugene Brevdo, and Honglak Lee.
\newblock Sample-efficient reinforcement learning with stochastic ensemble
  value expansion.
\newblock \emph{Conference on Neural Information Processing Systems}, 2019.

\bibitem[Kostrikov et~al.(2020{\natexlab{b}})Kostrikov, Yarats, and
  Fergus]{kostrikov2020image}
Ilya Kostrikov, Denis Yarats, and Rob Fergus.
\newblock Image augmentation is all you need: Regularizing deep reinforcement
  learning from pixels, 2020{\natexlab{b}}.

\bibitem[Nagabandi et~al.(2019)Nagabandi, Konolige, Levine, and
  Kumar]{pddm2019Nagabandi}
Anusha Nagabandi, Kurt Konolige, Sergey Levine, and Vikash Kumar.
\newblock Deep dynamics models for learning dexterous manipulation.
\newblock \emph{Conference on Robot Learning}, 2019.

\bibitem[Haarnoja et~al.(2018)Haarnoja, Zhou, Abbeel, and
  Levine]{haarnoja2018soft}
Tuomas Haarnoja, Aurick Zhou, Pieter Abbeel, and Sergey Levine.
\newblock Soft actor-critic: Off-policy maximum entropy deep reinforcement
  learning with a stochastic actor.
\newblock \emph{International Conference on Machine Learning}, 2018.

\end{thebibliography}
\bibliographystyle{unsrtnat}

\newpage
\appendix
\section{Theoretical Results}\label{app:theoretical_proofs}

We base our approach on the following theoretical results from the paper.

\begin{em}
{\bf Theorem 1 Restated.}
Consider a POMDP $\mdp$, and let $\latent_t$ be a latent space representation of the history and belief state such that $P(\state_t|\obs_{\leq t}, \action_{<t}) = P(\state_t|\latent_t)$. Let the policy class be such that $\action_t \sim \policy(\cdot | \latent_t)$, so that $P(\state_t|\latent_t, \action_t)=P(\state_t|\latent_t)$. Let $D_f$ be a generic $f-$divergence. Then the following inequalities hold:
$$D_f(\rho_\mdp^{\policy}(\obs, \action)||\rho_\mdp^E(\obs, \action))\leq D_f(\rho_\mdp^{\policy}(\state, \action)||\rho_\mdp^E(\state, \action))\leq D_f(\rho_\mdp^{\policy}(\latent, \action)||\rho_\mdp^E(\latent, \action))$$
\end{em}

\begin{proof}
The condition $P(\state_t|\latent_t, \action_t)=P(\state_t|\latent_t)$ essentially states that the actions of both the agent and the expert do not carry additional information about the state beyond what is available in the history. This will be true of all agents trained based on some representation of the history or just the current observation, and only excludes policies trained on ground truth states. Since we cannot hope to compete with policy classes that fundamentally have access to more information like the ground truth state, we believe this is a benign assumption.
%We should note that a similar assumption would be needed to prove the first inequality namely: $\mathcal{U}(\obs_t|\state_t, \action_t)=\mathcal{U}(\obs_t|\state_t)$, however this holds from the structure of the POMDP. 
With these assumptions, the proof is straightforward application of the data-processing inequality \cite{ali166datainequality}. 
%We will prove that $D_f(\rho_\mdp^{\policy}(\state, \action)||\rho_\mdp^E(\state, \action))\leq D_f(\rho_\mdp^{\policy}(\latent, \action)||\rho_\mdp^E(\latent, \action))$:

\begin{align}
D_f(\rho_\mdp^{\policy}(\latent, \action)||\rho_\mdp^E(\latent, \action)) & = \mathbb{E}_{\latent, \action\sim \rho_\mdp^E(\latent, \action)}\Bigg[f\Bigg(\frac{\rho_\mdp^{\policy}(\latent, \action)}{\rho_\mdp^E(\latent, \action)}\Bigg)\Bigg]\\
& = \mathbb{E}_{\latent, \action\sim \rho_\mdp^E(\latent, \action)} \mathbb{E}_{\state\sim P(\state|\latent)}\Bigg[f\Bigg(\frac{\rho_\mdp^{\policy}(\latent, \action)}{\rho_\mdp^E(\latent, \action)}\frac{P(\state|\latent)}{P(\state|\latent)}\Bigg)\Bigg] \\
& = \mathbb{E}_{\latent, \state, \action\sim \rho_\mdp^E(\latent, \state, \action)} \Bigg[f\Bigg(\frac{\rho_\mdp^{\policy}(\latent, \state, \action)}{\rho_\mdp^E(\latent, \state, \action)}\Bigg)\Bigg] \\
& = \mathbb{E}_{\state, \action\sim \rho_\mdp^E(\state, \action)} \Bigg[\mathbb{E}_{\latent\sim \rho_\mdp^{E}(\latent|\state, \action)}f\Bigg(\frac{\rho_\mdp^{\policy}(\latent, \state,  \action)}{\rho_\mdp^E(\latent, \state, \action)}\Bigg)\Bigg] \\
& \geq \mathbb{E}_{\state, \action\sim \rho_\mdp^E(\state, \action)} \Bigg[f\Bigg(\mathbb{E}_{\latent\sim \rho_\mdp^{E}(\latent|\state, \action)}\frac{\rho_\mdp^{\policy}(\latent, \state,  \action)}{\rho_\mdp^E(\latent, \state, \action)}\Bigg)\Bigg] \\
& = \mathbb{E}_{\state, \action\sim \rho_\mdp^E(\state, \action)} \Bigg[f\Bigg(\mathbb{E}_{\latent\sim \rho_\mdp^{E}(\latent|\state, \action)}\frac{\rho_\mdp^{\policy}(\state,  \action)\rho_\mdp^{\policy}(\latent|\state, \action)}{\rho_\mdp^E(\state, \action)\rho_\mdp^E(\latent|\state, \action)}\Bigg)\Bigg] \\
& = \mathbb{E}_{\state, \action\sim \rho_\mdp^E(\state, \action)} \Bigg[f\Bigg(\mathbb{E}_{\latent\sim \rho_\mdp^{\policy}(\latent|\state, \action)}\frac{\rho_\mdp^{\policy}(\state,  \action)}{\rho_\mdp^E(\state, \action)}\Bigg)\Bigg] \\
& = \mathbb{E}_{\state, \action\sim \rho_\mdp^E(\state, \action)} \Bigg[f\Bigg(\frac{\rho_\mdp^{\policy}(\state,  \action)}{\rho_\mdp^E(\state, \action)}\Bigg)\Bigg] \\
& = D_f(\rho_\mdp^{\policy}(\state, \action)||\rho_\mdp^E(\state, \action))
\end{align}

The first two equalities (9-10) follow from the fact that $\rho_\mdp^{\policy}(\state | \latent, \action) = P(\state|\latent) = \rho_\mdp^E(\state | \latent, \action)$ from the assumptions of the Theorem. The inequality (13) is a direct application of Jensen's inequality and the definition of an $f-$divergence. The other part of the main result follows the same reasoning, considering the observation model $\mathcal{U}(\obs|\state)$, rather than the belief distribution $P(\state|\latent)$. We also refer readers to \citet{gangwani2019learning} for similar derivations and analysis.
\end{proof}

\begin{em}
{\bf Lemma 1 Restated.}
Suppose we have an $\alpha$-approximate dynamics model given by $\mathbb{D}_{TV}(\widehat{\dynamics}(\state, \action), \dynamics(\state, \action)) \leq \alpha \ \forall (\state, \action)$. Let $R_{\max} = \max_{(s, a)} \mathcal{R}(\state,\action)$ be the maximum of the unknown reward in the MDP with unknown dynamics $\dynamics$. For any policy $\policy$, we can bound the sub-optimality with respect to the expert policy $\policy^E$ as:
\begin{equation*}
    \abs{J(\policy^E, \mdp) - J(\policy, \mdp) } \leq \frac{R_{\max}}{1-\gamma} \ \mathbb{D}_{TV}(\rho^\policy_{\mdphat}, \rho^E_\mdp) +  \frac{\alpha \cdot R_{\max}}{(1-\gamma)^2}.
\end{equation*}
\end{em}
\begin{proof}
The proof is a simple application of the triangle inequality on $\mathbb{D}_{TV}$. In particular we have:
\begin{align}
    \abs{J(\policy^E, \mdp) - J(\policy, \mdp) } & \leq \frac{R_{\max}}{1-\gamma} \ \mathbb{D}_{TV}(\rho^\policy_\mdp, \rho^E_\mdp) \\
    & \leq \frac{R_{\max}}{1-\gamma} \ \left( \mathbb{D}_{TV}(\rho^\policy_{\mdphat}, \rho^E_\mdp) + \mathbb{D}_{TV}(\rho^\policy_{\mdphat}, \rho^\policy_\mdp) \right) \\
    & \leq \frac{R_{\max}}{1-\gamma} \ \mathbb{D}_{TV}(\rho^\policy_{\mdphat}, \rho^E_\mdp) +  \frac{\alpha \cdot R_{\max}}{(1-\gamma)^2}.
\end{align}
Eq. (18) is directly from the definition, $(1-\gamma) \cdot J(\policy, \mdp) = \E_{(\state, \action) \sim \rho^\policy_\mdp} \left[ r(\state, \action) \right]$. Eq.~(19) is based on triangle inequality by decomposing with $\rho^\policy_{\mdphat}$ as an intermediate variable. The final inequality in Eq.~(20) is a direct consequence of error amplification lemma from \citet{RajeswaranGameMBRL}.
\end{proof}

\section{Practical Algorithm with Variational Models}\label{app:practical_VMAIL}

The divergence bound of Theorem \ref{thm:divergence_bound} allows us to develop a practical algorithm if we can learn a good belief state representation. Towards that end we turn to the theory of deep Bayesian filters \citep{karl2017deep} and begin with the likelihood:

$$    \log P(\obs_{1:T}|\action_{1:{T}}) = \log \int\prod_{t=1}^T \mathcal{U}(\obs_t|\state_t)\dynamics(\state_t|\action_{t-1}, \state_{t-1})d\state_{1:T}
$$
We can introduce the belief distribution $q(\latent_{1:T}|\obs_{1:T}, \action_{1:T-1})=\prod_{t=1}^T q(\latent_t|\obs_t, \latent_{t-1}, \action_{t-1})$, which considers only model classes that satisfy the the sufficient statistics requirement. Using the introduced belief distribution as the variational distribution, we derive the evidence lower bound (ELBO)~\citep{Blei2016VariationalIA, kingma2014autoencoding}:

$$    \log P(\obs_{1:T}|\action_{1:{T}}) = \log \int\prod_{t=1}^T \mathcal{U}(\obs_t|\latent_t)\dynamics(\latent_t|\action_{t-1}, \latent_{t-1})\frac{q(\latent_t|\obs_t, \latent_{t-1}, \action_{t-1})}{q(\latent_t|\obs_t, \latent_{t-1}, \action_{t-1})}d\latent_{1:T}
=$$

$$    \log \int q(\latent_{1:T}|\obs_{1:T}, \action_{1:T-1}) \prod_{t=1}^T \mathcal{U}(\obs_t|\latent_t)\frac{\dynamics(\latent_t|\action_{t-1}, \latent_{t-1})}{q(\latent_t|\obs_t, \latent_{t-1}, \action_{t-1})}d\latent_{1:T}
=$$

$$ \log \mathbb{E}_{q(\latent_{1:T}|\obs_{1:T}, \action_{1:T-1})}\left[\prod_{t=1}^T \mathcal{U}(\obs_t|\latent_t)\frac{\dynamics(\latent_t|\action_{t-1}, \latent_{t-1})}{q(\latent_t|\obs_t, \latent_{t-1}, \action_{t-1})}\right]    \geq $$

$$\mathbb{E}_{q(\latent_{1:T}|\obs_{1:T}, \action_{1:T-1})}\left[\log \prod_{t=1}^T \mathcal{U}(\obs_t|\latent_t)\frac{\dynamics(\latent_t|\action_{t-1}, \latent_{t-1})}{q(\latent_t|\obs_t, \latent_{t-1}, \action_{t-1})}\right]=
$$

$$\mathbb{E}_{q(\latent_{1:T}|\obs_{1:T}, \action_{1:T-1})}\left[ \sum_{t=1}^T \log \mathcal{U}(\obs_t|\latent_t)\right] + \mathbb{E}_{q(\latent_{1:T}|\obs_{1:T}, \action_{1:T-1})}\left[\sum_{t=1}^T\frac{\dynamics(\latent_t|\action_{t-1}, \latent_{t-1})}{q(\latent_t|\obs_t, \latent_{t-1}, \action_{t-1})}\right]=
$$

\begin{align*}
    & \mathbb{E}_{q(\latent_{1:T}|\obs_{1:T}, \action_{1:T-1})}\left[ \sum_{t=1}^T \log \mathcal{U}(\obs_t|\latent_t)\right] +  \\
    & \sum_{t=1}^T\mathbb{E}_{q(\latent_{1:t-1}|\obs_{1:t-1}, \action_{1:t-2})}\left[\mathbb{E}_{q(\latent_t|\obs_t, \latent_{t-1}, \action_{t-1})}\frac{\dynamics(\latent_t|\action_{t-1}, \latent_{t-1})}{q(\latent_t|\obs_t, \latent_{t-1}, \action_{t-1})}\right]=
\end{align*}

\begin{align*}
& \mathbb{E}_{q(\latent_{1:T}|\obs_{1:T}, \action_{1:T-1})}\left[ \sum_{t=1}^T \log \mathcal{U}(\obs_t|\latent_t)\right] + \\ 
& \sum_{t=1}^T\mathbb{E}_{q(\latent_{1:t-1}|\obs_{1:t-1}, \action_{1:t-2})}[\mathbb{D}_{KL}(q(\latent_{t}|\obs_{t}, \latent_{t-1}, \action_{t-1})||\dynamics(\latent_{t}|\latent_{t-1}, \action_{t-1}))]
\end{align*}

We represent $q, \dynamics$ and $\mathcal{U}$ as neural networks with parameters $\theta$. Following standard deep variational inference techniques \citet{kingma2014autoencoding} all distributions are represented as diagonal Gaussians parameterized by neural networks. To estimate the expectation, we can use sequential sampling from the belief distribution $\latent_{t}\sim q_{\theta}(\cdot|\obs_{t}, \latent_{t-1}, \action_{t-1}), t=1:T$ and the reparameterization trick~\cite{kingma2014autoencoding}. This leads to the empirical model loss:

\begin{equation} \label{eq:model_app}
\max_{\theta}\widehat{\mathbb{E}}_{q_{\theta}}\Big[
\sum_{t=1}^{T}\underbrace{\log\widehat{\mathcal{U}}_{\theta}(\obs_{t}|\latent_{t})}_{\text{reconstruction}}-\underbrace{\mathbb{D}_{KL}(q_{\theta}(\latent_{t}|\obs_{t}, \latent_{t-1}, \action_{t-1})||\widehat{\dynamics}_{\theta}(\latent_{t}|\latent_{t-1}, \action_{t-1}))}_{\text{forward model}}\Big].
\end{equation}

That is, we jointly train a belief representation $q_{\theta}$ and the Markovian dynamics model $\widehat{\dynamics}_{\theta}$, which allows us to optimize Eq. \ref{eq:our_objective_mdp} in our learned belief space. A number of recent works have considered similar models \citep{watter2015embed, zhang2019solar, SLAC20202Lee, gelada2019deepmdp, PlanNet2019Hafner, Dreamer2020Hafner}. We base our network architectural choice on the recurrent state space model \citep{PlanNet2019Hafner, Dreamer2020Hafner}, as it has shown strong performance in RL tasks from images. 

Once we learn a low-dimensional model, any on-policy RL algorithm can be used to train the policy using Eq.~\ref{eq:our_objective_pomdp}. In our setup, the RL objective is a differentiable function of the policy, model, and discriminator parameters. Based on this, we setup a $K$ step value expansion objective~\citep{feinberg2018modelbased, buckman2019sampleefficient} given below, and use it for policy learning.
\begin{equation} \label{eq:policy_app}
\max_{\policy_{\psi}}V_{\theta, \psi}^K(\latent_t)=\max_{\policy_{\psi}}\mathbb{E}_{\pi_\psi, \widehat{\dynamics}_\theta} \left[ \sum_{\tau=t}^{t+K-1}\gamma^{\tau-t}\log D_{\psi}(\latent^{\policy_{\psi}}_\tau, \action^{\policy_{\psi}}_\tau)+\gamma^{K}V_{\psi}(\latent^{\policy_{\psi}}_{t+K}) \right]    
\end{equation}

where the expectaion is taken over the sampling distributions $\action^{\policy_{\psi}}_t \sim \policy_{\psi}(\action|\latent_t^{\policy_{\psi}})$ and $\latent_{t+1}^{\policy_{\psi}} \sim \widehat{\dynamics}_{\theta}(\latent|\latent_t^{\policy_{\psi}}, \action^{\policy_{\psi}}_t)$. This allows us to optimize the policy $\policy_{\psi}$ by directly differenting the above objective through the learned model dynamics. The value function is fitted using the bootstrapped estimate in Eq. \ref{eq:policy_app}.

Finally, we train the discriminator $D_{\psi}$ in the learned latent space using Eq. \ref{eq:our_objective_mdp} with on-policy rollots from the model $\widehat{\dynamics}$:
\begin{equation}
    \label{eq:disc_app}
    \min_{D_\psi} \ \E_{(\latent, \action) \sim q_{\theta}(\rho_\mdp^E)} \left[ - \log D_\psi(\latent^E, \action^E) \right] \ + \ \E_{\latent^{\policy_{\psi}}, \action^{\policy_{\psi}}\sim\pi_\psi, \widehat{\dynamics}_\theta} \left[ - \log \left( 1 - D_\psi(\latent^{\policy_{\psi}}, \action^{\policy_{\psi}}) \right) \right]
\end{equation}

As outlined in Algorithm \ref{alg:vmail} we iteratively optimize the model, actor-critic and discriminator using Eq. \ref{eq:model_app}, \ref{eq:policy_app}, \ref{eq:disc_app}.

%%CF.5.24: well, but we need it to be efficient, so it's not just any policy optimiation algorithm -- it's important that it can optimize the policy entirely inside the model for efficiency.

%%CF.5.24: not explicitly clear that this is used in policy learning (i.e. as a terminal value function).

%%CF.5.24: do you use instance noise before passing things into the discriminator? (Can't remember if that was for this version of the method or an older version). If so, it's good to explain that and any other regularization here.

\section{Practical Off-Policy Imitation Learning Algorithms}\label{app:practical_offpolicy}
Training reinforcement learning policies from images is challenging using environment rewards, but even more so in the case of adversarial imitation learning. We explicitly choose to benchmark our method against SQIL and DAC, which use sample efficient off-policy training. In addition we can augment these approaches with state of the art method DrQ \citep{kostrikov2020image}, which has shown up to two orders of magnitude improvement in sample efficiency when training policies from raw pixels. The key of the DrQ approach is to introduce a family of image-augmentation functions $f(\state, v)$, where $\state$ is an environment state (a set of stacked images) and $v$ are augmentation parameters, from a fixed set of transformations. Given a batch of transition tuples $(\state_i, \action_i, \state'_i, \bm{r}_i)$ the standard Q-learning procedure is augmented as follows: the target values for the Bellman backups are computed as:

 \begin{equation}
     \bm{y}_i = \bm{r}_i+\gamma\frac{1}{K}\sum_{k=1}^K Q_{\theta}^{target}(f(\state_i', v'_{i,k}), \action'_{i,k}) \text{ where }  \action'_{i,k}\sim \policy(\cdot|f(\state_i', v'_{i,k}))
 \end{equation}
 
 while the Q-function is updated by:
 
 \begin{equation}\label{eq:drq}
    \theta\leftarrow\theta-\lambda\nabla_{\theta}\frac{1}{NM}\sum_{i=1, m=1}^{N, M}(Q_{\theta}(f(\state_i, v_{i, m}), \action_i)-\bm{y}_i)^2
  \end{equation}
 
 We can directly adapt SQIL to this setup, by using stationary rewards for the expert and policy replay buffers. For DAC, we train an additional discriminator $D_{\psi}$ minimizing the objective:
 
\begin{equation}
    \mathbb{E}_{\state, \action\sim\mathcal{B}^E}\Bigg[\frac{1}{K}\sum_{k=1}^K -\log D_{\psi}(f(\state, v_{k}), \action) \Bigg] + 
    \mathbb{E}_{\state, \action\sim\mathcal{B}^{\policy}}\Bigg[\frac{1}{K}\sum_{k=1}^K -\log (1-D_{\psi}(f(\state, v_{k}), \action)) \Bigg]
\end{equation}
 
 and then train the actor-critic algorithm with a modified version of Eq.  \ref{eq:drq}:

  \begin{equation}
     \bm{y}_i = \frac{1}{K}\sum_{k=1}^K \log D_{\psi}(f(\state_i, v_{i,k}), \action_{i}) + \gamma Q_{\theta}^{target}(f(\state_i', v'_{i,k}), \action'_{i,k})
 \end{equation}
 
 In our implementation the discriminator, critic and policy share the same convolutional encoder, which is trained using the discriminator and critic loss only. During training of this baseline, we noticed that periods of poor performance coincide with instability in the critic loss, rather than the discriminator. We hypothesise that this is potentially caused by issues with value function bootstrapping with non-stationary rewards or a mis-matched entropy objective from the soft actor-critic objective. We experimented with a number of mitigation strategies in an attempt to improve performance of this baseline, including constraining the discriminator, different regularization techniques, varying the buffer and batch sizes and separating the convolutional encoders of the discriminator and the actor/critic; however, these approaches didn't fully prevented the degradation in performance.

\section{Environments and Demonstration Data.}\label{app:data_environments}

\begin{figure}[t!]
  \begin{center}
    \includegraphics[width=0.195\textwidth]{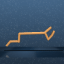}
    %\hspace{2mm}
    \includegraphics[width=0.195\textwidth]{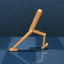}
    %\hspace{2mm}
    \includegraphics[width=0.195\textwidth]{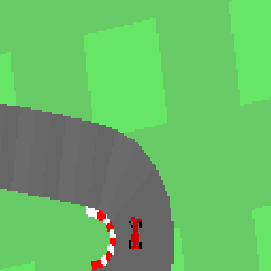}
    %\hspace{2mm}
    \includegraphics[width=0.195\textwidth]{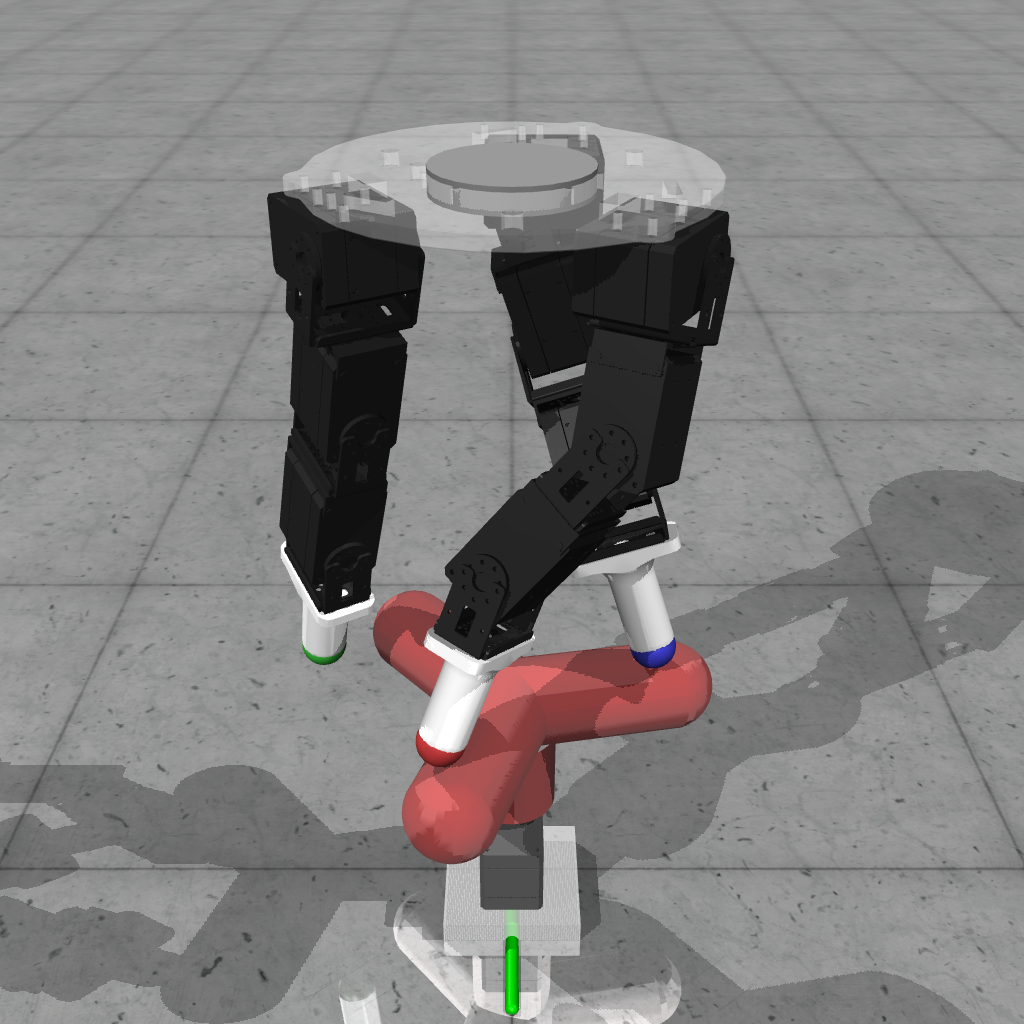}
    %\hspace{2mm}
    \includegraphics[width=0.195\textwidth]{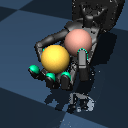}
    %\vspace{-4mm}
  \end{center}
  \caption{Illustration of the environments used in our experiments: Cheetah, Walker, Car Racing, D'Claw, and Baoding Balls. In all environments, the agent has access only to the RGB image frames as observations, except with additional access to proprioception in the Baoding Balls environment.}
\label{ref:envs}
\end{figure}

We consider the five visual control environments illustrated in Figure \ref{ref:envs}. The Cheetah Run and Walker Walk are standard tasks from the DeepMind Control Suite \citep{tassa2018deepmind}, with $64\times 64$ pixel observations. Following SQIL \cite{SQIL2020Reddy} we also consider the Car Racing environment from OpenAI Gym \cite{brockman2016openai}. We slightly modify the observation space, by removing the bottom part of the image, which contains episode and reward statistics and reshape the remaining image into $128\times128$ pixels. In addition, we benchmark our method on a custom D'Claw environment from the Robel suite \cite{Kumar_ROBEL}. The goal of the environment is to rotate the valve as fast as possible. We only use $128\times 128$ image observations without proprioception, which makes the task challenging due to a complex action and contact dynamics, as well as occlusions from the robot fingers. Our final environment is the Baoding balls task from \citet{pddm2019Nagabandi}. This is an extremely challenging task for policy learning, even in the state-based case. In addition to $128\times 128$ images, we also include proprioception information from the robot hand in the observation space. This is not unrealistic since the real ShadowHand robot platform can provide such information. 

In the main set of experiments, all methods receive access to 10 expert demonstrations, with the exception of the Baoding environment, which uses 25 demonstrations. The demonstrations for the DeepMind Control and D'Claw tasks are generated using a policy trained with SAC \cite{haarnoja2018soft}, the expert data for the Car Racing environment is generated using Dreamer \cite{Dreamer2020Hafner}, and the demonstrations for the Baoding task are generated using the PDDM framework \cite{pddm2019Nagabandi} from low-dimensional states.

We use three Walker-based locomotion environments from the DeepMind Control Suite \cite{tassa2018deepmind} for our transfer experiments. We train VMAIL on the Walker Stand and Walker Run tasks. The objective of the Walker Stand task is to maintain an upright position without moving, while the objective of the Run task is to maintain velocity greater than 8. We evaluate the transfer capabilities of VMAIL on Walker Walk, which aims to maintain velocity within the 2-4 range. We should note that these are qualitatively different behaviours. 
In a manipulation scenario, we use a set of custom D’Claw Screw tasks from the Robel suite \citet{Kumar_ROBEL}. The goal of all tasks is to rotate the valve as fast as possible in a particular direction. We train our model on tasks with a 3-prong valve with clockwise and counter-clockwise rotation, as well as a task with a 4-prong valve with counter-clockwise rotation.  We then evaluate the transfer to the task with a 4-prong valve with clockwise rotation. In all 4 environments we control the same 3-fingered D'Claw robot, but the shape of the valve object and the rotation direction vary. 
\end{document}